\documentclass[lettersize,journal,twoside]{IEEEtran}

\usepackage{amsmath,amsfonts,amssymb}
\usepackage{bm}

\usepackage{array}
\usepackage{textcomp}
\usepackage{verbatim}
\usepackage{ifthen}
\usepackage{enumitem}
\usepackage{pifont}
\usepackage{siunitx}

\usepackage[dvipsnames]{xcolor}
\usepackage{booktabs}
\usepackage{colortbl}

\usepackage{stfloats}
\usepackage{subcaption}
\usepackage[font=small]{caption}

\usepackage{tikz}
\usetikzlibrary{calc,math}

\usepackage{algorithm}
\usepackage[noEnd=true,indLines=true]{sty/algpseudocodex}

\usepackage[nocompress]{cite}

\usepackage[hidelinks]{hyperref}
\usepackage{cleveref}
\crefalias{ALG@line}{line}
\usepackage{orcidlink}

\makeatletter
\AddToHook{env/algorithmic/begin}{\def\@currentcounter{ALG@line}}
\makeatother

\usepackage{sty/custom}

\usepackage[dvipsnames]{xcolor}

\definecolor{linkcolor}{RGB}{0, 0, 255}

\title{Denoising Multirobot Trajectories}
\author{Yuhao Zhang, Keisuke Okumura, Ajay Shankar, and Amanda Prorok%
\thanks{Manuscript received: December 19, 2025; Revised: June 20, 2026; Accepted: September 11, 2026.
This work was supported in part by European Research Council (ERC) Project 949940 (gAIa), and by JST PRESTO (JPMJPR2513).
This paper was recommended for publication by Editor Jason O'Kane upon evaluation of the Associate Editor and Reviewers’ comments. \textit{(Corresponding author: Yuhao Zhang.)}}%
\thanks{The authors are with Department of Computer Science and Technology, University of Cambridge, U.K.
(e-mail: \texttt{yz981@cantab.ac.uk}; \texttt{\{ko393, as3233, asp45\}@cst.cam.ac.uk}).}%
\thanks{Keisuke Okumura is also with National Institute of Advanced Industrial Science and Technology (AIST), Japan.}%
\thanks{{Code and video}: \texttt{\textcolor{linkcolor}{https://github.com/proroklab/d4orm}}}%
}

\begin{document}

\bstctlcite{IEEEexample:BSTcontrol}

\markboth{IEEE Transactions on Robotics. Preprint Version. Accepted September, 2026}%
{Zhang \MakeLowercase{\textit{et al.}}: Denoising Multirobot Trajectories}

\IEEEpubid{%
  \parbox[t]{\textwidth}{%
    \footnotesize\centering
    \copyright~2026 IEEE. Personal use of this material is permitted.  Permission from IEEE must be obtained for all other uses, in any current or future media, including reprinting/republishing this material for advertising or promotional purposes, creating new collective works, for resale or redistribution to servers or lists, or reuse of any copyrighted component of this work in other works.
  }%
}

\maketitle

\begin{abstract}
Multi-robot trajectory planning is a fundamental problem in multi-robot coordination but remains computationally challenging due to its nonconvex, multimodal, and high-dimensional nature. This work builds upon \textsc{D4orm}, a dynamics-aware diffusion-denoising framework, and develops a family of planning architectures for diverse operational requirements. Unlike conventional numerical optimization methods, \textsc{D4orm} employs sampling-based optimization to generate solution trajectories through massively parallel sampling, leveraging modern computing architectures such as GPUs. Its diffusion-denoising structure iteratively optimizes \textit{deformations} to candidate control trajectories, providing an efficient and versatile paradigm for generating kinodynamically feasible and conflict-free trajectories. Using \textsc{D4orm} as the building block for advanced planners, we present a decoupled planner for improved scalability, an online receding-horizon planner with feedback control, and a distributed planner for resource-constrained settings. Evaluations with differential-drive and holonomic robots in 2D and 3D environments demonstrate that \textsc{D4orm}-based approaches find high-quality solutions faster and more reliably than other sampling-based optimization methods, such as MPPI, as well as a learned diffusion-model-based method. We further demonstrate zero-shot deployment on ten real quadrotors with obstacles, large-scale deconfliction with 100 simulated robots, and fully onboard distributed ``lifelong'' operation with six ground robots. Overall, these results establish diffusion denoising as a scalable and reliable framework for multi-robot coordination.
\end{abstract}

\section{Introduction}
\label{sec:introduction}

\IEEEPARstart{T}{eams} of robots are frequently required to occupy and operate in a shared workspace, and increasingly so in more modern applications. Settings such as those found in logistics, warehouse automation~\cite{wurman2008coordinating} and autonomous intersection management for vehicles~\cite{chen2015cooperative}, are all realistic, and present novel challenges for team-level autonomy and efficiency. Planning collision-free, kinodynamically feasible trajectories that respect the constraints of the entire team is thus fundamental to realizing such coordination. However, finding such a feasible solution is, in general, computationally intractable~\cite{hopcroft1984complexity,spirakis1984strong}. Furthermore, as these applications typically require planning with a limited time budget, sometimes involving a hundred robots or more, developing real-time and scalable \mbox{methodologies} presents one of the central challenges in the broader field of multi-robot research.

\input{figures/overview_v2}

Since a robot's motion can typically be represented as a continuous-time dynamical system, conventional approaches to trajectory planning can readily apply continuous \emph{numerical optimization} methods, such as gradient-based optimization or interior-point methods. These methods benefit from rich and mature mathematical tools with off-the-shelf solvers, providing practical solutions for synthesizing single-robot trajectories. However, the landscape is different for their multi-robot counterparts; three inter-related factors make it difficult to directly apply conventional numerical optimization methods:
\emph{(i)~nonconvexity}, arising from encoding inter-robot collision avoidance constraints,
\emph{(ii)~multimodality}, where numerous equally good solutions exist, along with many local but infeasible optima, and
\emph{(iii)~high dimensionality}, resulting from the rapidly growing joint solution space as the number of robots increases, which obscures useful information needed to derive a solution efficiently (e.g., gradient dilution).
Together, these factors pose significant computational challenges to deriving multi-robot trajectories reliably.

Consequently, recent research efforts have disfavored solving the problem in a fully joint, \emph{coupled} manner that considers the state space of the entire team simultaneously. Instead, decoupling the planning problem into subproblems for individual robots has become increasingly popular~\cite{luis2019trajectory,tordesillas2021mader,zhou2022swarm,csenbacslar2023rlss}. Such methods often build upon coupled trajectory optimization techniques while relaxing their formulations to reduce computational complexity and enhance tractability.~While this strategy exhibits better real-time responsiveness and scalability, decoupling prevents the discovery of tightly coordinated long-term behaviors due to the restricted solution space. In fact, the inability to coordinate beyond short-term collision avoidance can easily lead to deadlocks, livelocks, and congestion, which undermines the reliability of multi-robot systems.

\IEEEpubidadjcol
These limitations ultimately stem from the constraints of the underlying numerical optimization scheme. With a more expressive and flexible optimization framework tailored to multi-robot trajectories, solving multi-robot planning problems could become more tractable. This motivates us to explore \emph{sampling-based optimization} methods, which have increasingly attracted attention in challenging environments in recent years. This paradigm, of which a representative example is \emph{model predictive path integral control} (MPPI)~\cite{williams2018information}, employs a fundamentally different approach from conventional numerical optimization.
In particular, it is underpinned by principles of population-based optimization and aims to directly \emph{sample} solution trajectories through a large number of sampling attempts. This structure not only provides a practical solution to nonconvex optimization problems, but also benefits significantly from modern parallel computing architectures, including GPUs, culminating in real-time planning capabilities.

While sampling-based optimization has been actively applied to single-robot control~\cite{vlahov2024mppi}, its use for coupled multi-robot control remains under-explored. We attribute this primarily to the limited ability to extract informative search guidance for directing the sampling process in the very high-dimensional search spaces associated with multi-robot systems, which poses challenges analogous to those encountered in high-dimensional numerical optimization.

In parallel, recent advances in generative modeling within the machine learning community have produced highly capable methods, most notably \emph{diffusion models}~\cite{sohl2015deep,ho2020denoising}, which extract structural information from pure noise through iterative \emph{denoising} processes. Originally developed for image generation,  diffusion models have recently gained popularity as powerful tools for robot trajectory planning~\cite{janner2022planning,chi2023diffusion,carvalho2023motion} due to their ability to capture multimodal data structures in high-dimensional spaces. While many of these studies learn priors from demonstration trajectories, more recent works on trajectory optimization~\cite{pan2024model,xue2025full} draw a connection between diffusion-based denoising and sampling-based optimization, without any learning components. Although currently limited to single-robot applications, these studies demonstrate how diffusion structures can be leveraged to guide the sampling process and efficiently retrieve trajectory data in high-dimensional spaces, for example, in full-order locomotion behaviors.

{
\setlength{\tabcolsep}{3pt}
\renewcommand{\arraystretch}{1.2}
\newcommand{\rc}{\rowcolor[gray]{.9}}
\newcommand{\entry}[4]{
  \begin{minipage}[t]{0.04\linewidth} #2 \end{minipage} &
  \begin{minipage}[t]{0.1\linewidth} \textsc{#1} \end{minipage} &
  \begin{minipage}[t]{0.6\linewidth}#3\vspace{0.1cm}\end{minipage} &
  #4
}
\begin{table*}[ht!]
  \centering
  \small
  \caption{Summary of \textsc{D4orm} and its variants.}
  \label{table:summary}
  \begin{tabular}{lllll}
    \toprule
    \S && Description & Advantage
    \\
    \midrule
    \entry{D4orm~\cite{zhang2025d4orm}}{\ref{sec:preliminary:d4orm}}{
      The base sampling-based optimization method that uses a joint state representation.
    }{}
    \\
    \rc
    \entry{D4orm-D}{\ref{sec:d4orm_d}}{
      A semi-decoupled method that resolves conflicts iteratively for a subset of the team with \textsc{D4orm}.
    }{+Scalability}
    \\
    \entry{D4orm-C2F}{\ref{sec:d4orm_c2f}}{
      A receding-horizon framework that generates coarse references by \textsc{D4orm-D}, and tracks them with \textsc{D4orm} by generating fine trajectories.
    }{+Real-time responsiveness}
    \\
    \rc
    \entry{D-D4orm}{\ref{sec:d4orm_dist}}{
      Distributed version, allowing for onboard execution with inter-robot communication.%
    }{+Implementation flexibility}
    \\
    \bottomrule
  \end{tabular}
\end{table*}
}

Building upon these insights, combining sampling-based optimization with diffusion denoising offers a promising direction toward addressing key challenges in multi-robot trajectory planning, including nonconvexity, multimodality, and high dimensionality. Recent work \emph{\textsc{D4orm} (``deform'')}~\cite{zhang2025d4orm} provides a concrete instantiation of this idea through the optimization of \emph{\underline{d}ynamics-aware \underline{d}iffusion-\underline{d}enoised \underline{d}eformations}.
\textsc{D4orm} is an offline, centralized, coupled, full-horizon multi-robot trajectory optimizer that can accommodate arbitrary and mixed dynamical systems (e.g., double integrator, differential drive).
As shown in \cref{fig:overview}(a)--(c), it iteratively optimizes \emph{deformations} of the joint control trajectories for the entire team---incremental updates to the current solution candidate---where the optimization is realized through diffusion denoising without requiring analytical gradients.
Each denoising step is guided by an on-the-fly Monte Carlo gradient approximation that targets an interpretable optimization objective, and hence is also learning-free.
Furthermore, \textsc{D4orm} exhibits an \emph{anytime} property, with solutions improving over time, and it benefits greatly from GPU-based parallelization.

\textit{Contributions:}
In this work, we view this coupled multi-robot trajectory optimization method \textsc{D4orm} as a foundational building block for more advanced planning approaches and explore how it can be extended to more challenging settings.
We first analyze this method in detail, extending beyond~\cite{zhang2025d4orm}, and reveal its limitations, including a scalability bottleneck caused by gradient dilution as the number of robots increases (\cref{sec:d4orm-analysis}).
Next, as summarized in \cref{table:summary}, we introduce three variants of \textsc{D4orm}:
\emph{(i)}~a decoupled variant, \textsc{D4orm-D}, which iteratively applies \textsc{D4orm} to subsets of the entire team to enhance scalability (\cref{sec:d4orm_d});
\emph{(ii)}~an online receding-horizon variant, \textsc{D4orm-C2F}, which uses offline coarse trajectories generated by \textsc{D4orm-D} to guide online fine trajectory generation by \textsc{D4orm} (\cref{sec:d4orm_c2f}); and
\emph{(iii)}~a distributed version, \textsc{D-D4orm}, designed for resource-constrained environments, e.g., when GPUs are unavailable (\cref{sec:d4orm_dist}).
These extensions endow \textsc{D4orm} with desirable properties for trajectory planning---such as scalability, real-time responsiveness, and implementation flexibility---from demonstrations involving a 100-robot problem, to lifelong asynchronous operation in onboard distributed settings, as empirically shown in \cref{sec:evaluation2} and \cref{fig:overview}(e)--(f).
These results suggest that denoising multi-robot trajectories offers a new, practical, and promising instrument for solving various multi-robot planning problems.

The remainder of this paper presents the theoretical foundations and practical aspects of \textsc{D4orm}-based methodologies.
\cref{sec:related-work} situates \textsc{D4orm} within the context of multi-robot motion planning and planning with diffusion denoising.
\cref{sec:preliminary} provides the problem definition and introduces denoising optimization, specifically called \emph{model-based diffusion} (MBD)~\cite{pan2024model}, onto which \textsc{D4orm} builds, followed by \textsc{D4orm} itself in \cref{sec:preliminary:d4orm}.
\cref{sec:d4orm-analysis} onward contains the main contributions of this paper.

\textit{Difference from prior work:}
This paper substantially extends the work by Zhang et al.~\cite{zhang2025d4orm}, which introduces the basic \textsc{D4orm} framework.
Our work provides deeper insights into why iterative deformation with diffusion denoising successfully retrieves feasible, coordinated trajectories, going beyond other sampling-based optimization methods such as MPPI or MBD---a perspective that has received limited attention in prior work (\cref{sec:d4orm-analysis:theory}).
We further characterize the scalability limitations of vanilla \textsc{D4orm} in \cref{sec:d4orm-analysis:eval}, motivating the extensions developed in this paper.
With this improved understanding, we expand the denoising structure to support decoupled planning, receding-horizon planning, and distributed computation, which is amenable to distributed onboard deployment.
This comprehensive ecosystem, representing a significant departure from the prior work, opens up a new possibility for diffusion denoising as a powerful paradigm for multi-robot systems operating in traditionally hard-to-access high-dimensional spaces.

\section{Related Work}
\label{sec:related-work}

\subsection{Characteristics of Multi-Robot Planning}
There exists a rich body of literature on multi-robot trajectory planning.
To situate our contribution within the literature, we first outline a set of key axes that characterize methodologies in this domain.

\smallskip\noindent
\textbf{Planning Representation:}
\emph{Coupled} approaches plan over the joint state of all robots, solving a high-dimensional but globally consistent optimization problem.
In contrast, \emph{decoupled} approaches split the problem into robot-wise subproblems, which improves scalability but may sacrifice global optimality.

\smallskip\noindent
\textbf{When to Plan:}
\emph{Offline} methods complete planning prior to robot motion, which enables thorough coordination but lacks reactivity.
\emph{Online} methods repeatedly plan during execution, allowing adaptation to dynamic environments but under a limited planning time budget.

\smallskip\noindent
\textbf{Planning Horizon:}
\emph{Full-horizon} planning computes complete trajectories from start to goal in a single run.
By contrast, \emph{receding-horizon} approaches plan only a limited future horizon and replan continuously, enabling real-time execution but potentially sacrificing global foresight.

\smallskip\noindent
\textbf{Who Plans:}
\emph{Centralized} planning solves a single global problem assuming a central planner.
\emph{Distributed} approaches assign subproblems to multiple nodes.
\emph{Decentralized} methods are a special case of distributed planning, which let robots plan independently.
\emph{Offboard} planning assumes external computation, typically leveraging powerful compute infrastructure, while \emph{onboard} methods push computation to each robot, often under tighter resource constraints, but are more amenable to decentralized deployment.
\smallskip\noindent

We note that the original \textsc{D4orm} falls under \emph{coupled}, \emph{offline}, \emph{full-horizon}, and \emph{centralized offboard} planning.
This style can also serve as a core for developing alternative approaches such as scalable \emph{decoupled} methods, \emph{online receding-horizon} planning, and \emph{distributed} (potentially \emph{decentralized onboard}) planning, as summarized in \cref{table:summary}.
In terms of planning technique, it builds on \emph{sampling-based optimization}, but enhances this framework with techniques from \emph{diffusion} generative modeling, which we discuss next.

\subsection{Multi-Robot Trajectory Planning Methods}
The abstracted problem for multi-robot trajectory planning is known as \textit{multi-agent pathfinding} (MAPF)~\cite{stern2019def}, which aims to find collision-free time-parameterized paths on a graph under a discrete spatio-temporal representation, typically a four-connected 2D grid.
This abstraction naturally models warehouse automation~\cite{wurman2008coordinating}, one of the most successful applications of multi-robot systems.
Motivated by its industrial importance, a wide range of powerful algorithms has been developed.
In particular, recent offline, full-horizon, centralized methods~\cite{li2022mapf,okumura2024engineering} can solve instances with hundreds of agents almost instantly.
However, this abstraction does not generate kinodynamically feasible trajectories, limiting its ability to design smooth and agile coordination that can be directly executed by real robots.

To address trajectory planning with physical realism, two categories of approaches are commonly employed: \emph{sampling-based motion planning} (SBMP) and \emph{numerical optimization}.
The former, SBMP, applies discrete combinatorial search over a roadmap approximation of the joint continuous configuration space, constructed by random sampling~\cite{solovey2016finding,solis2021representation,kottinger2022conflict,okumura2023quick}.
The latter reduces multi-robot deconfliction to a numerical optimization problem and then typically applies well-established solvers to obtain a solution~\cite{augugliaro2012generation,kushleyev2013towards,adajania2023amswarm,tajbakhsh2024conflict}.
However, both categories suffer from a steep increase in computational burden as the number of robots grows, due to the curse of dimensionality.
This challenge has motivated a popular research thrust, particularly within numerical optimization methods, toward developing fully decoupled, receding-horizon trajectory planning approaches that ultimately aim to support decentralized onboard deployment~\cite{luis2019trajectory,tordesillas2021mader,zhou2022swarm,csenbacslar2023rlss}.
Nevertheless, such approaches remain prone to miscoordination issues such as local deadlocks, livelocks, and congestion, which can significantly degrade overall navigation throughput.

\textsc{D4orm} falls into a category orthogonal to SBMP and numerical optimization, which we call \emph{sampling-based optimization}.
These methods directly sample solutions from high-dimensional spaces through massive sampling attempts.
Unlike numerical optimization and SBMP, sampling-based optimization methods admit straightforward acceleration by parallelizing sampling attempts (e.g., with GPUs).
A notable example is \emph{model predictive path integral control} (MPPI)~\cite{williams2018information}, which has been successfully applied to single-robot control in challenging environments.
However, applying such methods to coupled multi-robot planning still faces the same dimensionality challenge.
In fact, existing sampling-based optimization methods for multi-robot setups rely on robot-wise decoupled representations with limited planning horizon~\cite{streichenberg2023multi,trevisan2024biased,jiang2024distributed,dergachev2025decentralized}, thereby sacrificing global coordination capability.
Moreover, even with such reduced planning spaces, these studies are generally limited to scenarios involving only about a dozen robots.
In contrast, we aim to sample joint trajectories directly through \emph{diffusion denoising}, a process that is better suited to handling high dimensionality, providing better scalability and real-time responsiveness.

\subsection{Planning with Diffusions}
Diffusion models, which first gained considerable attention in image generation~\cite{sohl2015deep,ho2020denoising,rombach2022high}, have recently emerged as a popular choice for learning-based robot motion generation~\cite{janner2022planning,chi2023diffusion,carvalho2023motion}.
A key strength of diffusion models lies in their ability to capture structure in very high-dimensional spaces.
They have also been applied in multi-agent contexts, such as trajectory prediction~\cite{jiang2023motiondiffuser} and motion planning with constrained optimization~\cite{shaoul2025multi,liang2025simultaneous,liang2025discrete}.
While these diffusion-based generative methods primarily aim to imitate demonstration trajectories, recent studies propose learning-free trajectory generation via Monte Carlo gradient approximation~\cite{pan2024model,xue2025full}, which bears a strong connection to MPPI, as discussed later in \cref{sec:d4orm-analysis}.
In particular, \emph{model-based diffusion} (MBD)~\cite{pan2024model}, described in detail in \cref{sec:preliminary:mbd}, was originally designed for single-robot planning.
\textsc{D4orm} extends it to multi-robot settings by introducing team-level cost functions and a key methodological innovation: the iterative optimization of \emph{deformation}, defined below, to solve more challenging problems.
This results in a capable optimization tool to design multi-robot coordination, which can also be extended to various multi-robot planning styles.

\section{Preliminaries}
\label{sec:preliminary}
In this section, we first define the multi-robot trajectory planning problem and provide a high-level illustration of the diffusion generative model.
We then explain the single-robot trajectory optimization method called \emph{model-based diffusion} (MBD)~\cite{pan2024model}, followed by its multi-robot extension \textsc{D4orm}.
Throughout this paper, a braced superscript (e.g., $\{k\}$) is used to index robots.

\subsection{Problem Definition}
\label{sec:preliminary:problem_definition}
The system comprises a team $R$ of $K$ spherical-shaped robots, with each robot having a radius of $\mathrm{R}_a \in \mathbb{R}_{>0}$.
The motion of each robot is governed by the dynamical system, $\dot{x} = f_{\mathrm{dyn}}(x, u)$, where the state is denoted by $x \in \mathcal{X} \subset \mathbb{R}^{d_x}$ and the control input by $u \in \mathcal{U} \subset \mathbb{R}^{d_u}$.
Unless otherwise stated, we assume that the team consists of homogeneous robots ($\mathrm{R}_a$ and $f_{\mathrm{dyn}}$ are the same for all robots), although the \textsc{D4orm} methods apply to heterogeneous cases.

Given joint initial and goal configurations $\mathcal{S}, \mathcal{T} \in \mathcal{X}^K$, the objective is to compute a \emph{solution} $\tau$, a list of trajectories for all $K$ robots, ensuring each trajectory is collision-free and dynamically feasible.
Specifically, the trajectory $\tau^{\{k\}} \in (\mathcal{X}, \mathcal{U})^H$ for each robot $k \in R$ spans a finite horizon of $H$ discrete time steps, with each step sampled at a fixed time interval $\Delta t \in \mathbb{R}_{>0}$.
Then, the trajectories need to satisfy the following:
\begin{alignat}{2}
    \tau^{\{k\}}[t+1] &= \texttt{RK4}(\tau^{\{k\}}[t], f_{\mathrm{dyn}}, \Delta t)
      &&\quad \text{(feasibility)} \nonumber \\
    \tau^{\{k\}}[1].x &= \mathcal{S}^{\{k\}}
      &&\quad \text{(initial state)} \nonumber \\
    \tau^{\{k\}}[H].x &= \mathcal{T}^{\{k\}}
      &&\quad \text{(goal state)} \nonumber \\
    \text{Dist}(\tau^{\{k\}}[t] &, \tau^{\{l\}}[t]) > 2\cdot\mathrm{R}_a
      &&\quad \text{(safety)}
    \label{eqn:problem-conditions}
\end{alignat}
Here, $\text{Dist}(\cdot)$ denotes the Euclidean distance between the positional components of two robot states, and $\texttt{RK4}$ refers to the classical fourth-order Runge–Kutta integration method applied to the system dynamics.

The quality of a feasible solution $\tau$, i.e., the reward to be maximized, is evaluated in relation to the total travel time, represented in the following form:
\begin{equation}
\frac{1}{KH}
  \sum_{k=1}^{K}
  \sum_{t=1}^{H}
   \mathbb{I}\left[ \tau^{\{k\}}[t].x = \mathcal{T}^{\{k\}} \right]
\label{eq:objective}
\end{equation}
where $\mathbb{I}[\cdot] = 1$ if the condition is true; zero otherwise.

Note that the full-state trajectory $\tau \in (\mathcal{X}^H)^K$, which represents a solution, is uniquely determined from a control trajectory $U \in (\mathcal{U}^H)^K$ for the team (i.e., a sequence of control commands) by rolling out the robots' states using $f_{\mathrm{dyn}}$, starting from the initial configuration $\mathcal{S} \in \mathcal{X}^K$.
Therefore, the optimization problem is, in fact, cast as optimizing for $U$ rather than for $\tau$.
Given a solution candidate $U$, we denote a \emph{deformation} to it with $\Delta U$, which has the same dimension as $U$.
We seek a new control sequence, $U + \Delta U$, that produces a full-state trajectory with a better objective score.

\subsection{Diffusion Generative Models}
\label{sec:preliminary:diffusion}
Let $p_0$ denote the target distribution of data $\tau$, which may represent images, robot trajectories, or other high-dimensional structured data.
\emph{Diffusion models}~\cite{ho2020denoising,song2020score} are a class of probabilistic generative models designed to transform simple noise distributions into the target $p_0$.
They operate through two Markov chains: \emph{(i)}~a forward process that gradually corrupts the data by adding noise, and \emph{(ii)}~a reverse denoising process that iteratively removes noise to recover data samples.
Mathematically, starting from a noise-free sample $\tau_0 \sim p_0$, Gaussian noise is sequentially added according to the schedule parameters $\alpha_i$, producing increasingly noisy samples $\tau_i$ and ultimately an isotropic Gaussian distribution $p_N$:
\begin{align}
  p_{i|i-1}(\tau_i \mid \tau_{i-1}) = \mathcal{N}\left(\sqrt{\alpha_i}\tau_{i-1}, (1 - \alpha_i) I\right).
\end{align}
The backward process $p_{i-1|i}(\cdot)$ then reverses the forward process $p_{i|i-1}(\cdot)$.
When this backward distribution is available, one can reconstruct $p_0$ through:
\begin{align}
  p_{i-1}(\tau_{i-1}) &= \int p_{i-1|i} (\tau_{i-1} | \tau_{i}) p_i(\tau_i) \, d\tau_i, \\
  p_0(\tau_0) &= \int p_N(\tau_N) \prod_{i=N}^1 p_{i-1|i}(\tau_{i-1} | \tau_i) \, d\tau_{1:N}.
\end{align}
With the score-based formulation under reverse-time stochastic differential equation evolution~\cite{song2020score}, this backward process can be approximated by iteratively applying:
\begin{align}
  \tau_{i-1} = \frac{1}{\sqrt{\alpha_i}}
  \left(
  \tau_i + \frac{1 - \alpha_i}{2}\nabla_{\tau_i}\log{p_i(\tau_i)}
  \right)
  + \sqrt{1-\alpha_i}\,\epsilon
  \label{eq:score-based-guidance}
\end{align}
where $\epsilon \sim \mathcal{N}(0, I)$~\cite{chan2024tutorial}.
Since the score function $\nabla_{\tau_i}\log{p_i(\tau_i)}$ is generally intractable, it is commonly approximated by a neural network trained to predict it.

\subsection{Trajectory Optimization with Diffusion Denoising}
\label{sec:preliminary:mbd}
Unlike conventional diffusion models, \emph{model-based diffusion} (MBD)~\cite{pan2024model} performs the backward process in an entirely learning-free and data-free manner to solve \emph{single-robot} trajectory optimization problems.
The core idea of MBD is to cast trajectory optimization as the problem of sampling a high-reward trajectory.

For now, we focus on the single-robot case.
Let $r(\tau)$ denote a reward function that evaluates the quality of a trajectory $\tau$.
For example, it may encode whether the robot reaches its goal efficiently while avoiding collisions.
Using a temperature parameter $\lambda \in \mathbb{R}_{>0}$, we define the target distribution $p_0$ as:
\begin{align}
  p_0(\tau) \propto \exp\left(\frac{r(\tau)}{\lambda}\right).
  \label{eq:p_0}
\end{align}
In this formulation, $p_0$ assigns higher probability density to trajectories with higher rewards.
Thus, the problem of finding optimal trajectory reduces to sampling trajectories from $p_0$.

Since the exact form of $p_0$ is generally intractable, we adopt generative modeling to optimize trajectories.
MBD addresses this by attempting to sample trajectories from $p_0$ through progressive denoising, akin to standard diffusion models.
Instead of relying on neural networks, MBD uses \emph{Monte Carlo score ascent}, which approximates the score function as follows:
\begin{align}
  \nabla_{\tau_i} \log p_i(\tau_i) \approx
  -\frac{\tau_i}{1 - \bar{\alpha}_i}
  + \frac{\sqrt{\bar{\alpha}_i}}{1 - \bar{\alpha}_i} \bar{\tau}_{0 \mid i}
  \label{eq:guidance}
\end{align}
where $\bar{\alpha}_i = \prod_{j = 1}^i \alpha_j$.
The term $\bar{\tau}_{0 \mid i}$ denotes a reward-weighted average trajectory, defined as
\begin{align}
  \bar{\tau}_{0 \mid i}
  =
  \frac{\sum_{\tau \in \Gamma} p_0(\tau) \tau}{\sum_{\tau \in \Gamma} p_0(\tau)}
  \approx
  \sum\nolimits_{\tau \in \Gamma}
  \frac
      {\exp\left( \frac{r(\tau)}{\lambda} \right)}
      {\sum\nolimits_{\tau' \in \Gamma} \exp\left( \frac{r(\tau')}{\lambda} \right)} \, \tau
  \label{eq:mc}
\end{align}
where $\Gamma$ is a set of trajectories sampled from a Gaussian distribution parameterized by $\tau_i$ and $\bar{\alpha}_i$:
\begin{equation}
  \Gamma \sim \mathcal{N}\left( \frac{\tau_i}{\sqrt{\bar{\alpha}_i}}, \left( \frac{1}{\bar{\alpha}_i} - 1 \right) I \right).
\end{equation}
Then, one can obtain noise-free trajectories using \cref{eq:score-based-guidance}, with the approximated score function from \cref{eq:guidance}.

MBD further accelerates the denoising process by replacing \cref{eq:score-based-guidance} with
\begin{align}
  \tau_{i-1} = \frac{1}{\sqrt{\alpha_i}} \left( \tau_i + (1 - \bar{\alpha}_i) \nabla_{\tau_i} \log p_i(\tau_i) \right),
  \label{eq:denoising}
\end{align}
where the score term is given by \cref{eq:guidance}.

An important issue not yet addressed is that a trajectory must be kinodynamically feasible, which requires states and controls to be consistent with the given dynamics $f_{\mathrm{dyn}}$.
Thus, sampling states and controls independently is impractical, as the distribution would collapse to a Dirac delta function.
Therefore, similar to MPPI, MBD samples control trajectories instead, and then recovers state–control trajectories by rolling out from the initial state, i.e., iteratively generating a state sequence with $x[t+1] = \texttt{RK4}(x[t], u[t], f_{\mathrm{dyn}}, \Delta t)$ given a control sequence.
This rollout process over the trajectory batch $\Gamma$ can be efficiently parallelized (e.g., on GPUs).

{
\begin{algorithm}[t!]
\caption{Denoising Optimization (MBD)~\cite{pan2024model}}
\label{algo:mbd}
\begin{algorithmic}[1]
\small
\State $U_N \sim \mathcal{N}(\bm{0}, I)$
\label{algo:mbd:init}
\Comment{initialize control sequence}
\For{$i \gets N~\text{to}~1$}
\State Sample $M$ control sequences:
\begin{align*}
\Gamma_u \sim
  \mathcal{N} \left( \frac{U_i}{\sqrt{\bar{\alpha}_i}}, \left( \frac{1}{\bar{\alpha}_i} - 1 \right) I \right)
\end{align*}\;
\label{algo:mbd:sample}
\State Obtain $M$ trajectories:
$\Gamma \leftarrow \texttt{rollout}\left(\mathcal{S}, \Gamma_u \right)$
\label{algo:mbd:rollout}
\State Compute Monte Carlo estimation:
\begin{align*}
\bar{U} \leftarrow \frac
{\sum_{\tau \in \Gamma} p_0(\tau) (\tau.u)}
{\sum_{\tau \in \Gamma} p_0(\tau)}, \:
p_0(\tau) \approx \exp\left(\frac{r(\tau)}{\lambda}\right)
\end{align*}
\label{algo:mbd:mc}
\State Estimate Score:
$
\nabla_{U_i}\log p_i(\tau_i)
\approx
-\frac{U_i}{1-\bar{\alpha}_i}
+
\frac{\sqrt{\bar{\alpha}_i}}{1-\bar{\alpha}_i}\,\bar{U}.
$
\label{algo:mbd:estimate}
\State Perform one-step denoising:
\begin{align*}
U_{i-1}
=
\frac{1}{\sqrt{\alpha_i}}
\left(
U_i
+
(1-\bar{\alpha}_i)\nabla_{U_i}\log p_i(\tau_i)
\right)
\end{align*}
\label{algo:mbd:denoising}
\EndFor
\State \Return $\texttt{rollout}\left( \mathcal{S}, U_0 \right)$
\label{algo:mbd:final}
\end{algorithmic}
\end{algorithm}
}

The resultant trajectory optimization scheme by MBD, referred to herein as the \emph{denoising optimization}, is summarized in \cref{algo:mbd}.
The process begins with an initial noisy control sequence $U_N \in \mathcal{U}^H$, sampled from a Gaussian distribution (\cref{algo:mbd:init}).
Based on this noisy trajectory, $M$ control samples are generated (\cref{algo:mbd:sample}).
These samples are then rolled out into trajectories (\cref{algo:mbd:rollout}) and used to approximate the score (\cref{algo:mbd:estimate}).
An update is subsequently applied using \cref{eq:denoising} (\cref{algo:mbd:denoising}).
These operations are repeated until the number of denoising steps $N$ is reached, and the final denoised control sequence is returned as a solution (\cref{algo:mbd:final}).
To improve stability during the denoising optimization process, rewards are normalized across the batch when computing the score~\cite{ioffe2015batch}.
Note that enforcing control constraints (actuator limits) can be achieved by action clamping during trajectory rollout, as commonly employed in MPPI implementations~\cite{williams2018information}.

\subsection{\textsc{D4orm} for Multi-Robot Trajectories}
\label{sec:preliminary:d4orm}
{
\begin{algorithm}[t!]
  \caption{Iterative Deformations (\textsc{D4orm})~\cite{zhang2025d4orm}}
\label{algo:d4orm}
\begin{algorithmic}[1]
\small
\State Initialize $\tau$
\While{not interrupted}
\State Get a deformation control $\Delta U$ for fixed $\tau.U$ using \cref{algo:mbd}
\State $\tau \leftarrow \texttt{rollout}\left( \mathcal{S}, \tau.U + \Delta U \right)$
\EndWhile
\State \Return $\tau$
\end{algorithmic}
\end{algorithm}
}

We now provide an outline of \textsc{D4orm}\cite{zhang2025d4orm} as the base coupled and full-horizon method that our work subsequently builds upon.
\textsc{D4orm} extends the denoising optimization (\cref{algo:mbd}) to multi-robot settings by introducing team-level fitness functions (i.e., reward) and, as a key methodological innovation, iteratively optimizing \emph{deformation} vectors to address more challenging problems.

In \textsc{D4orm}, the single-robot control sequence $U_i \in \mathcal{U}^{H}$ becomes a joint control sequence $U_i \in (\mathcal{U}^{H})^K$ that includes all $K$ robots.
The rollout procedure in \cref{algo:mbd} is then applied to this joint control sequence to compute the joint state sequence, yielding a joint trajectory of $K$ robots, denoted as $\tau \in \mathbb{R}^{H \times K(d_x + d_u)}$.
The denoising process operates on this joint representation to obtain a solution $\tau$ that optimizes a specified reward $r(\tau)$.

Consistent with the objective function \cref{eq:objective},
a reward for the
joint trajectory $\tau$ can now be defined as
\begin{align}
r(\tau) = \frac{1}{KH} \sum_{k = 1}^K  \sum_{t=1}^{H} \left( r_\text{goal}(\tau^{\{k\}}, t) + r_\text{safe}(\tau^{\{k\}}, t) \right).
\label{eq:generic-reward}
\end{align}
The goal reward $r_\text{goal}$ encourages robot \( k \) to approach its target position \( p_\mathcal{T}^{\{k\}} \) as early as possible, while the safety term $r_\text{safe}$ penalizes collisions with other robots:
\begin{equation}
r_{\text{goal}}(\tau^{\{k\}}, t) = 1 - \frac{\| p^{\{k\}}[t] - p_\mathcal{T}^{\{k\}} \|}{\| p^{\{k\}}[1] - p_\mathcal{T}^{\{k\}} \|}
\end{equation}
\begin{equation}
r_{\text{safe}}(\tau^{\{k\}}, t) =
    \begin{cases}
        -1,  & \text{if } \| p^{\{k\}}[t] - p^{\{l\}}[t] \| \leq 2\mathrm{R}_a + \epsilon \\
        0,   & \text{otherwise}
    \end{cases}
\label{eq:reward_safe}
\end{equation}
where \( \epsilon \in \mathbb{R}_{>0} \) defines a safety margin, and \( l \in R \setminus \{k\} \).

This extension to a multi-robot setting is non-trivial in practice due to the larger scale and complexity of the problem, which makes selecting a preset number of denoising steps ($N$) difficult.
To address this, \textsc{D4orm} (\cref{algo:d4orm}) applies the denoising optimization of \cref{algo:mbd} iteratively.
Instead of na\"{i}vely optimizing for a new control sequence from scratch each iteration, \textsc{D4orm} applies \cref{algo:mbd} with a small fixed $N$ to optimize a \emph{deformation} $\Delta U \in (\mathcal{U}^H)^K$ that improves a candidate solution trajectory.
The samples in \cref{algo:mbd} now correspond to \emph{deformations} ($\Delta U_\text{sample}$) that will be applied to the control sequence from the previous iteration (the candidate solution $\tau.U$), and the set of sampled state trajectories is obtained by rolling out $\tau.U + \Delta U_\text{sample}$.
The candidate solution trajectory is updated as $\tau.U \leftarrow \tau.U + \Delta U$ only after a full run of \cref{algo:mbd} that yields the optimized $\Delta U$.
In other words, rather than directly modeling $p_0$ of the final trajectory with MBD in a single pass, \textsc{D4orm} iteratively models the \emph{deformations} to accommodate problems with varying complexity commonly encountered in multi-robot trajectory planning.

Note that since each iteration refines $\tau$ closer towards the optimal trajectory, \textsc{D4orm} is an \textit{anytime} algorithm (Line 2, \cref{algo:d4orm}).
The termination criteria may vary based on the application: it is possible to stop as early as the first successful solution is found, or continue until some computational budget is exhausted.
Our work utilizes this flexibility of \textsc{D4orm}.
Nevertheless, as we will show in \cref{sec:d4orm-analysis}, using $\Delta U$ as the optimization target is not merely a desirable feature, but is in fact, instrumental to its correctness.
Moreover, in contrast to standard diffusion models, which rely on data-driven neural networks and suffer from distribution shifts across changing environments and robot dynamics, \textsc{D4orm} naturally adapts to this through its model-based formulation.

\section{Analysis of \textsc{D4orm}}
\label{sec:d4orm-analysis}
While \textsc{D4orm}'s optimization process yields a powerful trajectory optimization framework, as we will show, it has several limitations.
Our objective in this work is to build upon the original algorithm, referred to as ``vanilla" or ``base" \textsc{D4orm} in this text, and broaden its applicability to more complex planning settings.
Particularly in contrast with prior work, we are interested in tackling environments that contain obstacles as well as heterogeneous teams of robots.
As such, we modify the reward term $r_\text{safe}$ defined in \cref{eq:reward_safe} to assign an additional penalty of $-1$ whenever the corresponding state $\tau^{\{k\}}[t]$ involves a collision with the environment, or more generally, violates a state constraint.
Heterogeneous settings are accommodated by performing rollouts under the respective dynamical systems $f^{\{k\}}_\text{dyn}$ for each robot and by evaluating collisions based on robot-specific radii $\mathrm{R}^{\{k\}}_a$.

To unravel the difficulties in developing such extensions, we begin by establishing a deeper analysis of \textsc{D4orm}'s characteristics.
We first derive its mathematical connections to existing sampling-based methods such as MPPI and MBD, and provide new qualitative and quantitative evaluations on typical multi-robot trajectory planning problems.
We also analyze the scalability limitations of \textsc{D4orm}, revealing a failure mode caused by gradient dilution and thereby motivating the development of more advanced methods.

\subsection{Mathematical Connections to MPPI and MBD}
\label{sec:d4orm-analysis:theory}

We now explain why \textsc{D4orm} is particularly well-suited for solving multi-robot trajectory planning, by clarifying its mathematical foundations in comparison with MPPI and MBD.
{In short, \textsc{D4orm} uses the deformation-optimizing style of MPPI, but with diffusion-inspired noise annealing from MBD.}

Before proceeding, we clarify the terminology of \textit{step} and \textit{iteration}. A \textit{step} refers to a single sampling process followed by an update process. An \textit{iteration} refers to a complete MBD optimization procedure consisting of $N$ denoising steps.

\subsubsection{MPPI to MBD}
\label{sec:mppi_to_mbd}
A popular sampling-based optimization method for trajectory planning is MPPI~\cite{williams2018information}, which computes the optimal control sequence $U^\ast$ according to:
\begin{align}
  U^\ast \leftarrow \frac{1}{\eta}\sum_{\tau \in \Gamma}
  {\exp{\left(\frac{r(\tau)}{\lambda} + \mathcal{I}(\tau) \right)}} (\tau.U),
  \label{eq:mppi-vanilla}
\end{align}
where $\Gamma$ denotes a set of state–control trajectories obtained by rolling out control samples drawn from $\mathcal{N}(\hat{U}, \Sigma)$, with a reference control trajectory $\hat{U}$, a pre-specified noise parameter $\Sigma$, and a normalization constant $\eta$.
Here, $\mathcal{I}(\tau)$ represents an importance-sampling weight derived from an information-theoretic perspective, although it is often disabled in practice due to stability concerns~\cite{vlahov2024mppi}.
By removing the $\mathcal{I}$ term, \cref{eq:mppi-vanilla} reduces to:
\begin{align}
  U^\ast &\leftarrow
  \frac{1}{\eta}
  \sum_{\tau \in \Gamma}
  \exp{\left(\frac{r(\tau)}{\lambda} \right)} (\tau.U),
  \label{eq:mppi-softmax}
\end{align}
which is equivalent to MBD when $N = 1$;
in other words, an MPPI update corresponds to a single denoising step, as discussed in~\cite{xue2025full}.

A key distinction from MPPI, however, lies in the use of noise annealing in MBD, governed by the schedule of $\alpha_i$, which facilitates convergence toward (locally) optimal solutions.
\cref{fig:1d_opt} illustrates the behavioral differences between MBD and MPPI on a synthetic, highly non-convex one-dimensional objective function with multiple local minima.
By introducing a noise schedule, MBD achieves faster convergence, as the covariance of the sampling distribution $\left( \frac{1}{\bar{\alpha}_i} - 1 \right) I$ decreases as $i{\rightarrow}1$.
This encourages wide exploration in the early stages and focused local search in the later, less-noisy stages for exploitation.
In contrast, MPPI relies on a predefined and fixed covariance matrix for the sampling distribution; consequently, it may exhibit fluctuating behavior across steps when the covariance is too large.
This motivates the use of MBD over vanilla MPPI in \textsc{D4orm}, particularly since multi-robot trajectory planning inherently involves highly multimodal, nonconvex optimization problems.

From an optimization perspective, MPPI can also be viewed as a multi-step deformation-optimizing algorithm. At each step, control trajectories are sampled from $\mathcal{N}(U_i,\Sigma)$, which is equivalent to sampling perturbations $\Delta U \sim \mathcal{N}(\mathbf{0},\Sigma)$ and applying them to the current solution $U_i$. Each update therefore estimates a deformation $\Delta U^*$ of the current trajectory.

In contrast, MBD cannot be directly interpreted as a deformation-optimizing method. Its sampling distribution is
$
\mathcal{N}\left(
\frac{U_i}{\sqrt{\bar{\alpha}_i}},
\left(\frac{1}{\bar{\alpha}_i}-1\right)I
\right),
$
which means that perturbations are applied to the scaled trajectory $\frac{U_i}{\sqrt{\bar{\alpha}_i}}$ that depends on the current denoising step index, rather than directly to $U_i$. Such an optimization process is therefore coupled to the diffusion schedule and cannot be naturally decoupled into a sequence of deformation updates. As a result, MBD is inherently formulated as a single-iteration algorithm consisting of a predefined sequence of denoising steps. This motivates the development of \textsc{D4orm}, which instead performs iterative deformation optimization across multiple iterations.

\input{figures/1d_opt}

\subsubsection{MPPI to \textsc{D4orm}}
While MBD is formulated as a single-iteration algorithm that directly optimizes a control trajectory, we reinterpret the optimization process as solving for a deformation that can be applied iteratively. This leads to \textsc{D4orm}. Compared with MPPI, where each step computes a deformation through a single softmax update, \textsc{D4orm} computes a deformation using an entire MBD diffusion-denoising process. In other words, MPPI performs one step per deformation update, whereas \textsc{D4orm} performs a complete MBD optimization to obtain each deformation before applying it to the current trajectory estimate.

\subsubsection{MBD to \textsc{D4orm}}
\label{sec:mbd-d4orm}
The iterative deformation, absent in MBD, provides practical benefits for multi-robot trajectory planning.
As shown in \cref{fig:1d_opt} (top right), MBD can still become trapped in local optima regardless of the number of denoising steps.
Consequently, a single denoising optimization may yield infeasible solutions, as demonstrated in \cref{fig:overview}(b) for the 10-robot 2D case.
The iterative denoising in \textsc{D4orm} alleviates this issue by resetting the noise schedule at each iteration to re-enable exploration, while the optimization objective remains the deformation vector.
In other words, \textsc{D4orm} instantiates multiple runs of MBD sequentially, with each iteration optimizing a deformation over the result of previous iteration.

It may be possible to use MBD iteratively by initializing a variable $U_N$ at \cref{algo:mbd:init} in \cref{algo:mbd} with the previously optimized result (i.e., $U_0$).
However, this does not provide an effective iterative refinement scheme, as it violates MBD’s assumption that $U_N$ is sampled from the normal distribution $\mathcal{N}(0,I)$, leading to a mismatch in the designed noise annealing;
i.e., iterative MBD causes a distribution shift issue, especially the mean of the sampling distribution at first denoising step is further scaled by $\frac{1}{\sqrt{\bar{\alpha}_N}}$ (\cref{algo:mbd:sample} in \cref{algo:mbd}), hence iterative attempts do not improve the solution trajectory.
Meanwhile, \textsc{D4orm} overcomes this issue, as it always optimizes deformation from a normal distribution, and each denoising attempt preserves MBD’s assumption.
This structural difference will be empirically validated in \cref{sec:d4orm-analysis:evals:base}.

\subsubsection{MPPI vs. MBD vs. \textsc{D4orm}}
In summary, MPPI can be interpreted as a multi-step deformation-optimizing algorithm, whereas MBD is a single-iteration optimization algorithm. \textsc{D4orm} unifies these two perspectives by performing iterative deformation optimization across multiple iterations. Consequently, when each iteration consists of exactly one step, \textsc{D4orm} reduces to MPPI. When there is only one iteration, \textsc{D4orm} becomes equivalent to MBD.

\subsection{Empirical Evaluation of \textsc{D4orm}}
\label{sec:d4orm-analysis:eval}
We now establish the performance of the base \textsc{D4orm} framework qualitatively and quantitatively using commonly employed metrics similar to prior work~\cite{zhang2025d4orm}.
We extend the analysis to now consider obstacles and heterogeneous teams, and also compare it against a joint-space numerical optimization routine as well as a method based on learned diffusion models.
We use the following kinodynamic models:
\begin{itemize}[leftmargin=*]
\item \textbf{Differential Drive} system that represents wheeled robots with state $x = [p_x, p_y, \theta, v]^\top$, control $u = [\omega, a]^\top$, and $f_{\mathrm{dyn}}(x, u) = [v\cos\theta, v\sin\theta,\omega, a]^T$;
\item \textbf{2D Holonomic} double-integrator system that represents ground robots with state $x=[p_x, p_y, v_x, v_y]^\top$, control $u=[a_x, a_y]^\top$, and $f_{\mathrm{dyn}}(x, u) = [v_x, v_y, a_x, a_y]^T$;
\item \textbf{3D Holonomic} double-integrator system, which is a 3D version of the 2D case;
\item \textbf{Heterogeneous} scenario including half differential-drive and half holonomic robots, with robot max acceleration set to different values.
\end{itemize}
We refer to scenarios with circular obstacles as \textbf{2D Holo X Obs}, where X denotes the number of obstacles.
Visual illustration of such an instance is available in \cref{fig:d4orm_c2f_procedure}.
All evaluations are carried out on a laptop PC with an Intel Core i9-13900HX CPU, equipped with an NVIDIA RTX 4080 GPU.
Our implementation is based on the accompanying code of MBD~\cite{pan2024model} and is implemented in Python using the \texttt{JAX} library~\cite{jax2018github} to streamline GPU acceleration.
The planning horizon is set to $H{=}100$, with $M{=}1{,}024$ trajectory rollouts used to approximate gradients.

\subsubsection{Qualitative Analysis}
\cref{fig:overview}(a)--(c) presents a qualitative investigation of the intermediate steps of the denoising process for each system.
\cref{fig:overview}(c) considers a navigation task involving eight heterogeneous robots, initially positioned around a circle, with their respective goals located at antipodal positions.
Such scenarios are widely used as benchmarks in the multi-robot planning literature, as they inevitably force planners to make non-trivial deconfliction efforts.
\cref{fig:overview}(b) illustrates the refinement across denoising steps, i.e., over $i$ in \cref{algo:mbd}, as well as iterations of \cref{algo:d4orm}.
The figure shows that trajectories converge toward their targets while remaining collision-free as the denoising steps increase, despite the enormous number of optimization variables, i.e., $d_u \times H \times K$.
In particular, \cref{fig:overview}(a) highlights how sampling attempts converge to high-quality solutions due to the noise annealing mechanism, an effect not observed in MPPI alone, as confirmed in \cref{fig:1d_opt}.~As seen in the 2D holonomic case with 2,000 variables in \cref{fig:overview}(b), iterative deformation optimization compensates for incomplete trajectories and further improves solution quality by escaping local minima through resetting the diffusion noise scheduler.

The generated trajectories by \textsc{D4orm} can be deployed on real robots in a zero-shot manner.
\cref{fig:overview}(d) presents a deployment snapshot of a team of ten quadrotors~\cite{woo2025sanity} operating in the presence of obstacles.

\subsubsection{Comparison with Sampling-Based Optimization}
\label{sec:d4orm-analysis:evals:base}
Next, we compare \textsc{D4orm} with MPPI and MBD to gain empirical insight into how it differs from these sampling-based optimization methods.
Following the strategy described in \cref{sec:mbd-d4orm}, MBD is evaluated using its iterative variant denoted as i-MBD, which directly optimizes the trajectory initialized from the previous iteration, rather than optimizing trajectory deformations as in \textsc{D4orm}.
For evaluation, these methods are adapted to perform sampling in the joint control space.
The rollout implementation is shared across methods, while the optimization update rules are implemented separately and efficiently in JAX.
We use a temperature parameter of $\lambda{=}0.3$, which is hand-tuned in the 2D holonomic environment and kept fixed across all other environments.
We evaluate three key metrics of interest:
\emph{(i)~runtime}, to quantify scalability against varying team sizes by measuring clock time for computing successful solutions,
\emph{(ii)~success rate}, to quantify the probability of finding a solution within a deadline, and
\emph{(iii)~reward}, to quantify solution quality according to \cref{eq:objective} when planning is allowed to continue for longer.

\input{figures/baseline_cmp}

\cref{fig:results} shows a complete overview of our comparisons over 50 runs.
First, compared to MPPI, we observe that \textsc{D4orm} achieves a noticeable improvement in both runtime and success rate for \emph{all} robot models and for \emph{all} team sizes.
In particular, the success rate plots show that for the 2D holonomic case, \textsc{D4orm} can always retrieve kinodynamically feasible and collision-free trajectories for 16-robot team in obstacle-free environments given a \SI{3}{\second} deadline.
The average runtime is lower, $\approx$\SI{1.25}{s}, which is a near $2\times$ improvement over MPPI.
The reward plots show the ``anytime planning'' nature using the number of steps, which serves as a metric for assessing runtime independently of computing environments.
Recall that the reward is an indicator for solution quality, and as such, both methods converge to similar reward values given sufficient time for refinement.
However, \textsc{D4orm} shows two key advancements.
First, its reward curves show a steeper gradient during early steps, indicating that we obtain high-quality solutions faster.
Second, as indicated by the solid dots on the curves, \textsc{D4orm}'s high-quality solutions are successful (i.e., conflict-free) earlier than MPPI of similar quality.
These improvements are owing to the ability of diffusion denoising to capture complicated, multimodal reward distributions in high-dimensional spaces.
These findings remain consistent even in advanced scenarios, such as those involving obstacles and heterogeneous cases.
An empirical comparison with \emph{cross-entropy method} (CEM)~\cite{botev2013cross}, another representative sampling-based optimization method, is provided in \cref{fig:baseline_cmp_appendix} of the Appendix and exhibits trends similar to those observed in the comparison between \textsc{D4orm} and MPPI.

The second primary observation is that, as discussed in \cref{sec:mbd-d4orm}, the iterative variant of MBD struggles to solve challenging deconfliction instances even with additional iterations, due to distribution shift.
Indeed, as shown in the success rate and reward plots, increasing the number of iterations does not necessarily improve performance.
Meanwhile, \textsc{D4orm} improves the success rate over iterations, presenting its structural advantage of optimizing deformation instead of directly optimizing control variables.
We also provide a comparison with vanilla MBD (single-iteration with 10,000 steps) in \cref{fig:baseline_cmp_appendix} of the Appendix to demonstrate that a predefined number of denoising steps does not generalize efficiently across problems with different levels of complexity.

\subsubsection{Comparison with Numerical Optimization Method}
\label{sec:d4orm-analysis:evals:ipopt}
{
\newcommand{\e}[1]{{\scriptsize \m{\pm#1}}}
\setlength{\tabcolsep}{5pt}
\begin{table}[t!]
\centering
\caption{Runtime [\SI{}{\second}] of two methods to find feasible solutions in 2D holonomic case.}
\label{table:d4orm_vs_numerical}
\small
\begin{tabular}{rrrrr}
\toprule
\#robots & 4 & 8 & 12 & 16 \\ \midrule
\texttt{IPOPT} & 1.12\e{0.526} & 6.33\e{2.284} & 35.32\e{11.53} & failed \\
\textsc{D4orm} & 0.12\e{0.002} & 0.13\e{0.002} & 0.30\e{0.217} & 1.25\e{1.016}
\\
\bottomrule
\end{tabular}
\end{table}
}

We next contrast \textsc{D4orm} with a naive numerical optimization approach.
Specifically, we encode \cref{eqn:problem-conditions} using a direct collocation method~\cite{kelly2017introduction} and solve it with the seminal off-the-shelf nonlinear optimization solver \texttt{IPOPT}~\cite{wachter2006implementation}, one of the most widely used approaches for trajectory optimization. Our \texttt{IPOPT} implementation first solves a warm-start trajectory that ignores collisions. The resulting solution is perturbed with small random noise, which we found improves solution diversity and reduces average computation time, and then used to initialize a collision-aware nonlinear optimization that refines the trajectories into a feasible deconflicted solution.
We carefully align the objective functions and constraints implementation of both approaches, ensuring the comparison is conducted as fairly as possible.

The results on the 2D holonomic case are summarized in \cref{table:d4orm_vs_numerical}, which highlight the high efficiency of \textsc{D4orm} compared to the standard numerical optimization approach.
As the number of robots increases, the numerical optimization exhibits a steep rise in runtime, whereas \textsc{D4orm} maintains modest planning times.
This is inherently due to the highly parallel structure of sampling-based optimization, which benefits from GPU acceleration.

\subsubsection{Comparison with Learned Diffusion Method}
\label{sec:d4orm-analysis:evals:mmd}
\input{figures/vs_mmd}
We further compare \textsc{D4orm} against a method based on learned diffusion models. Notably, training diffusion models directly for multi-robot systems is challenging due to distribution shifts induced by changes in team size.
Therefore, an existing coupled strategy called \textit{multi-robot multi-model planning with diffusion} (MMD)~\cite{shaoul2025multi} employs a learned diffusion model to generate collision-aware trajectories for individual robots, while identifying and resolving inter-robot conflicts using combinatorial search methods developed for discrete MAPF, such as \textit{conflict-based search} (CBS)~\cite{sharon2015conflict}.
While this design effectively mitigates the distribution shift associated with changing team sizes,
it comes at the cost of significantly increased planning times due to the repeated interaction between diffusion-model inference and search-based conflict management.
Indeed, as shown in \cref{fig:vs_mmd}, \textsc{D4orm} achieves substantially lower runtime while generating smoother trajectories, highlighting the efficiency of sampling-based optimization and iterative deformation framework for centralized coupled planning.
We additionally note that MMD requires substantial effort before the actual planning phase, as it relies on training models tailored to the target robot platform.
This requirement further limits its applicability across different robot dynamics.

\subsection{Limitations: Performance and Scale}
Although \textsc{D4orm}'s iterative refinement is highly effective in solving multi-robot trajectory planning, its performance degrades in settings with even larger team sizes.
This limitation arises from two main reasons.

First, as the number of robots increases, \textsc{D4orm} may allow a small number of robots to collide in order to maximize the overall team reward defined by \cref{eq:generic-reward}, which it may prioritize increasing the goal reward $r_\text{goal}$ at the cost of reducing the safety reward $r_\text{safe}$.
This occurs because \textsc{D4orm} optimizes trajectories to maximize the \textit{total} team reward without imposing strict constraints to avoid collisions.
If only a few robots are involved in collisions, their negative contribution to the total reward is small, reducing the planner’s incentive to resolve these conflicts.
Simply increasing the penalty associated with collisions does not help, as it can lead to overly conservative behavior in which robots avoid the risk of collision by failing to progress toward their goals.

\input{figures/d4orm_scalability2}

Second, the gradient information diminishes with increasing robots.
\cref{fig:d4orm_scalability} illustrates this phenomenon, which shows trajectories computed by \textsc{D4orm}, while ignoring collisions (i.e., dropping $r_\text{safe}$).
The task here is to find a robot-wise trajectory from point to point but in a joint manner.
From a single-robot trajectory optimization perspective, the difficulty of the problem does not vary regardless of how many robots are involved.
However, since \textsc{D4orm} performs a \emph{joint} optimization, the reward gradient weakens with a large number of optimization variables involved.
The consequence, shown in \cref{fig:d4orm_scalability} (top), is a degradation of the trajectory quality obtained after the first iteration.
The decrease in normalized reward for one iteration, as well as the number of iterations required to derive goal-reaching trajectories in \cref{fig:d4orm_scalability} (bottom), also provide evidence of this issue.

These inherent limitations of vanilla \textsc{D4orm} motivate us to develop its advanced versions to pursue scalable methods, which we introduce in the following sections.

\section{\textsc{D4orm-D}: Failure-Aware Decoupling}
\label{sec:d4orm_d}
In the previous section, we discussed two key limitations of vanilla \textsc{D4orm}, particularly when scaling to many robots:
\emph{(i)}~some robots tend to prioritize reaching their goals over avoiding collisions, and
\emph{(ii)}~the gradient computed by Monte Carlo ascent becomes weaker, leading to slower convergence.
A practical strategy to address these issues is to \emph{decouple} the planning representation from the full joint optimization problem.
This section introduces \textsc{D4orm-D}, which iteratively refines the trajectories of subsets of the team to effectively repair infeasible solutions, inspired by an advanced MAPF method~\cite{li2022mapf}.

{
\begin{algorithm}[t]
  \caption{\textsc{D4orm-D}: Failure-Aware Decoupling}
\label{algo:d4orm_d}
\begin{algorithmic}[1]
\small
\State $\tau \gets$ \textsc{D4orm} without considering inter-robot collisions
\While{not interrupted}
\State $\F \gets$ groups of failed robots in $\tau$ if exists; or $\{R\}$
\For{$\G \in \F$}
\State Compute deformation $\Delta U^{\{\G\}}$ using \cref{algo:mbd}
\State $\tau^{\{ \G \}} \leftarrow \texttt{rollout}\left( \mathcal{S}^{\{ \G \}}, \tau^{\{ \G \}}.U + \Delta U^{\{\G\}} \right)$
\EndFor
\EndWhile

\State \Return $\tau$

\end{algorithmic}
\end{algorithm}
}

\cref{algo:d4orm_d} presents the pseudocode for \textsc{D4orm-D}, which repeatedly applies \textsc{D4orm} to \emph{failed} robots.
Given a potentially infeasible solution $\tau$, we define a robot $k$ as failed if $\tau^{\{k\}}$ involves collisions or does not reach its goal.
All failed robots are then grouped as $\F = \{ \G_1, \G_2, \ldots \}$ based on mutual collisions:
\begin{align}
  \texttt{collide}(i, j)\quad &\forall i, j \in \G_k, i \neq j,
  \\
  \lnot\texttt{collide}(i, j)\quad &\forall i \in \G_k, j \in \G_l, k \neq l,
\end{align}
where \texttt{collide} denotes whether two trajectories in $\tau$ collide.
Robots that only collide with obstacles or fail to reach their goals are assigned to singleton groups.
This grouping is achieved by constructing an adjacency graph over robots based on collisions and extracting its connected components. We also augment each group using collision information from the previous iteration. Specifically, if a robot from a previous collision group is currently colliding with a robot outside that group, we reconnect it to its previous direct neighbors that are collision-free in the current iteration, as these robots are more likely to collide again and therefore benefit from joint optimization. The final collision groups are then defined as the connected components of this augmented graph.

Once \F is identified, we apply \cref{algo:mbd} to compute \emph{deformations} for each group \G.
Each group’s trajectory is refined according to the rewards of its own robots only, without being influenced by rewards from other groups.
\textsc{D4orm-D} repeats this identification and repair operation until a user-specified termination condition is met.
Implementation-wise, once a feasible solution is obtained, it is stored, and subsequent iterations continue with fully joint optimization to further refine the solution, while the algorithm can still return the stored feasible solution at any time.

\begin{figure}[t!]
  \centering
  \newcommand{\entry}[2]{
  \begin{tikzpicture}
  \node[](img) at (0, 0) {
  \includegraphics[
    width=0.21\columnwidth,
    trim={2cm 1.8cm 1.2cm 1.1cm},
    clip
  ]{images/d4orm_d_procedure/decouple_#1.pdf}\hfill
  };
  \node[anchor=north west] at (img.north west) {\scriptsize (#2)};
  \end{tikzpicture}%
  }
  \entry{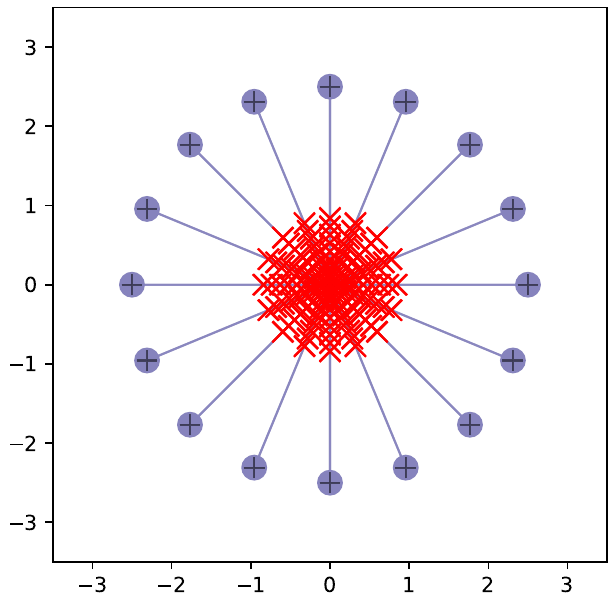}{a}%
  \entry{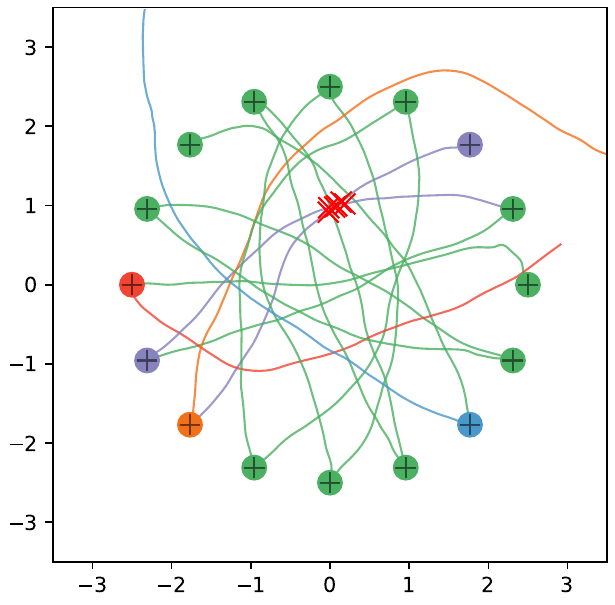}{b}%
  \entry{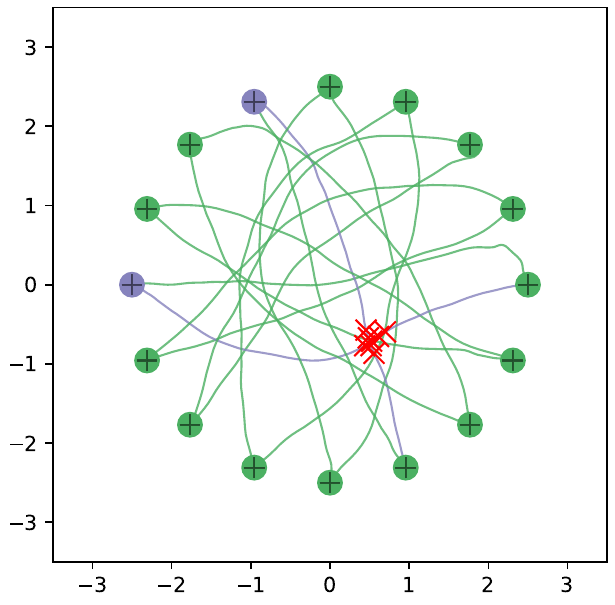}{c}%
  \entry{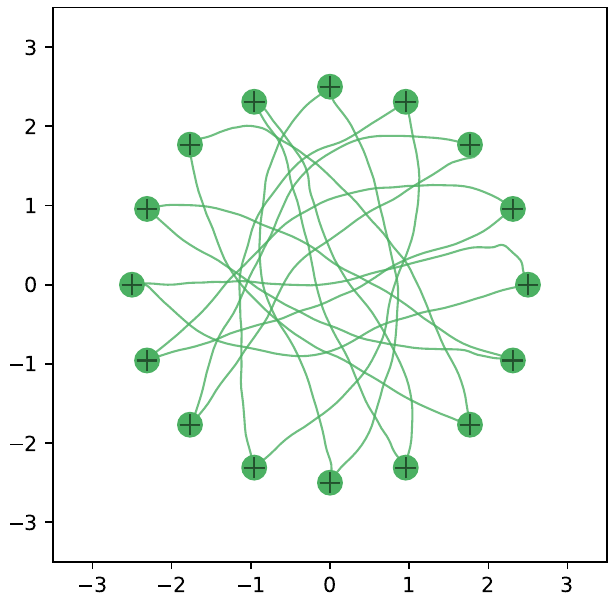}{d}%

  \caption{Visualization of \textsc{D4orm-D} trajectory generation over iterations.
  \textcolor{red}{$\times$} marks the step where a collision occurs.
  \textcolor[HTML]{2ca02c}{$\bullet$} indicates successful robots, other colors indicate failed robots belonging to the same group.}
  \label{fig:d4orm_d_procedure}
\end{figure}

\cref{fig:d4orm_d_procedure} illustrates an example run of \textsc{D4orm-D}.
In the first phase (a), we perform \textsc{D4orm} with just one iteration, yielding a single group that includes all robots, since they all pass through the center of the circle at the same time step.
Subsequent refinements are then performed for each group in $\F$ using \textsc{D4orm}, with successful robots treated as dynamic obstacles.
In this example, this decoupled replanning eventually produces collision-free trajectories in (d).

The rationale of \textsc{D4orm-D} is that decoupling the team to isolate failed robots allows the denoising optimization to focus effectively on resolving their failures.
Collision-based grouping naturally divides a large team into smaller subgroups, enabling more accurate reward-gradient approximation by Monte Carlo ascent.
Moreover, iterative decoupling enables robots from different groups to be dynamically reassigned for joint optimization.
Finally, soft constraints (i.e., collision penalty) adopted in the reward calculation facilitate the replanning of previously successful robots, enhancing the planner’s ability to escape highly constrained infeasible solutions.

\section{\textsc{D4orm-C2F}: Coarse-to-Fine Planning}
\label{sec:d4orm_c2f}

Although failure-aware decoupling offers greater capability in generating successful trajectories, it still encounters two primary limitations.
First, decoupling speeds up \textsc{D4orm}'s planning, but generating a successful trajectory for several robots still remains computationally intensive, enough to prevent real-time planning.
Second, the approach operates in an open-loop manner, making it susceptible to disturbances and uncertainties in the actual environment and dynamics model, which limits its effectiveness in real-world scenarios.

To address these challenges, we propose a coarse-to-fine planning strategy.
This approach first generates a full-horizon but \emph{coarse} joint trajectory, which reduces computational overhead and admits a few potential collisions.
Then, online planning is conducted along the coarse trajectory to generate an executable \emph{fine} trajectory while dynamically avoiding collisions in a closed-loop manner.
In this scheme, the coarse trajectory serves as a form of ``guidance,'' embedding long-term coordinated behaviors such as congestion mitigation~\cite{kato2025congestion} for a short-term, fine-grained, and reactive trajectory planner.
\cref{algo:d4orm_c2f} encapsulates the idea, termed \textsc{D4orm-C2F}, which is described in detail below.
\cref{fig:d4orm_c2f_procedure} illustrates its running example.

\subsection{Coarse Trajectory Generation}
The most computationally expensive component in the denoising optimization is the rollout.
This involves simulating the control sequence over a horizon $H$ (e.g., set to 100 in \cref{sec:d4orm-analysis:eval}) and computing the corresponding reward.
Since rolling out a single trajectory is a sequential process, reducing the horizon length $H$ can significantly improve efficiency.

To exploit this insight, we first use \textsc{D4orm-D} to compute a coarse trajectory for each robot.
We reduce the rollout horizon and proportionally increase the time step $\Delta t$ as $H\coarse{=}H / f\coarse$ and $\Delta t\coarse{=}\Delta t \cdot f\coarse$, where $f\coarse$ is a parameter that controls the horizon length.
In practice, $f\coarse$ is set to a small integer (e.g., set to 2 in \cref{sec:evaluation2}) so as not to introduce significant deviations from the actual solutions.

To further reduce the overhead of computing the coarse trajectory, \textsc{D4orm-C2F} lowers the resolution of collision checking when identifying failed robots in \textsc{D4orm-D} by performing checks only every $o\coarse$ steps.
Even with this sparse checking, all steps are still optimized under the original reward structure, thereby implicitly encouraging collision avoidance.
Compared to alternative heuristics, such as relaxing collision constraints or reducing the weight of $r_\text{safe}$, this approach relaxes only the success criterion for the coarse trajectory, which we found to provide effective guidance.

With these relaxations, \textsc{D4orm-D} generates a coarse trajectory $\tau\coarse$ with $H\coarse$ steps (\cref{algo:d4orm_c2f:d4orm-d}), where every $o\coarse$ step is guaranteed to be collision-free, as shown in \cref{fig:d4orm_c2f_procedure}(b).
We then interpolate $\tau\coarse.x$ to obtain $p\coarse$, see \cref{fig:d4orm_c2f_procedure}(a), an $H$-step path that specifies robot positions with a time step of $\Delta t$ (\cref{algo:d4orm_c2f:interpolate}).
Although the unverified steps may still contain collisions, generating such coarse guidance provides a useful initial plan that can be refined during online execution. The impact of the choices of $f\coarse$ and $o\coarse$ is analyzed in \cref{tab:d4orm_c2f_parameter_tab} of the Appendix.

{
\begin{algorithm}[t!]
\caption{\textsc{D4orm-C2F}: Coarse-to-Fine Planning}
\label{algo:d4orm_c2f}
\begin{algorithmic}[1]
\small
\item[\textbf{Parameters:}]
\Statex $H, \Delta t$ \Comment{Target planning horizon and time step}
\Statex $H\predict, H\execute$ \Comment{Local predict and execute horizon}
\Statex $f\coarse$ \Comment{Scaling factor for coarse time step}
\Statex $o\coarse$ \Comment{Offset for coarse collision checking}
\Statex $m$ \Comment{Index margin for subgoal selection}
\medskip
\Statex {\small \emph{(Coarse trajectory preparation)}}
\State $H\coarse \gets H / f\coarse$;\; $\Delta t\coarse \gets \Delta t \cdot f\coarse$
\State $\tau\coarse \gets$ \textsc{D4orm-D} with $(H\coarse, \, \Delta t\coarse, o\coarse)$
\label{algo:d4orm_c2f:d4orm-d}
\State $p\coarse \gets \texttt{interpolate}(\tau\coarse.x, H)$
\label{algo:d4orm_c2f:interpolate}
\State Refine $\tau\coarse[\frac{j+m}{f\coarse}:]$ with $(H\coarse,\Delta t\coarse,o{=}1)$ in the backend using latest available $j$ and update $p\coarse$
\label{algo:d4orm_c2f:refine}
\medskip
\Statex {\small \emph{(Online fine trajectory generation and execution)}}
\While{not all robots reach goal}
\label{algo:d4orm_c2f:online-start}
%
\State $j \gets$ ave. of the nearest subgoal indices in $p\coarse$ for $R$
\State $p\subgoal \gets p\coarse[j + m]$
\label{algo:d4orm_c2f:subgoal}
\State $\tau\predict \gets$ \textsc{D4orm} with $(H\predict, \, \Delta t, \, p\subgoal)$
\label{algo:d4orm_c2f:d4orm}
\State Execute $\tau\predict[1, 2, \ldots, H\execute].U$
\label{algo:d4orm_c2f:execute}
\EndWhile
\label{algo:d4orm_c2f:online-end}
\end{algorithmic}
\end{algorithm}
}

{
\newcommand{\entry}[2]{
  \begin{minipage}[b]{0.156\linewidth}
    \centering
    \includegraphics[width=1\linewidth,clip,trim={2cm 1.5cm 1.2cm 1.2cm}]
                    {images/d4orm_c2f_procedure/#1}
    {\scriptsize #2}
  \end{minipage}
}

\begin{figure*}[t!]
  \centering
  \entry{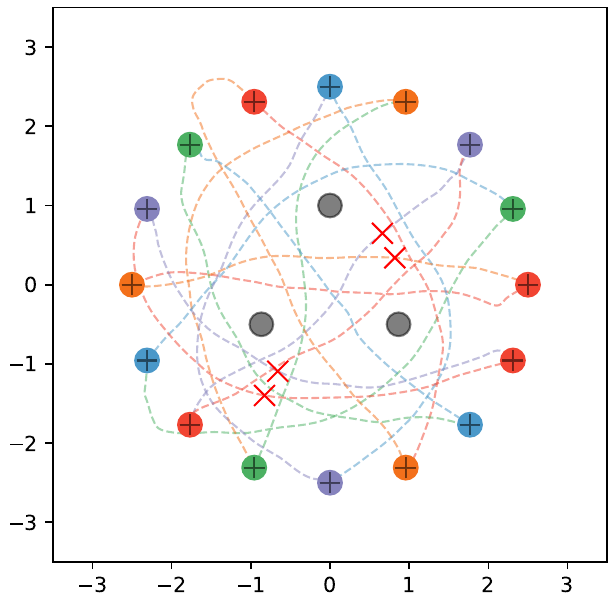}
        {(a) Initial Coarse Path}%
  \entry{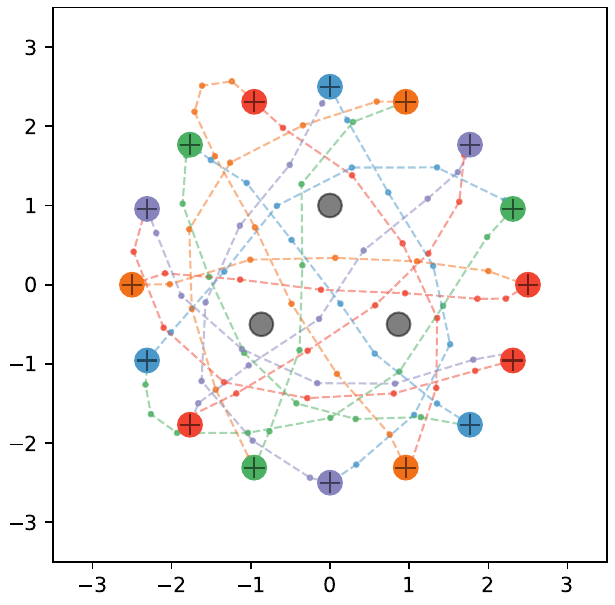}
        {(b) Verified Waypoints}%
  \entry{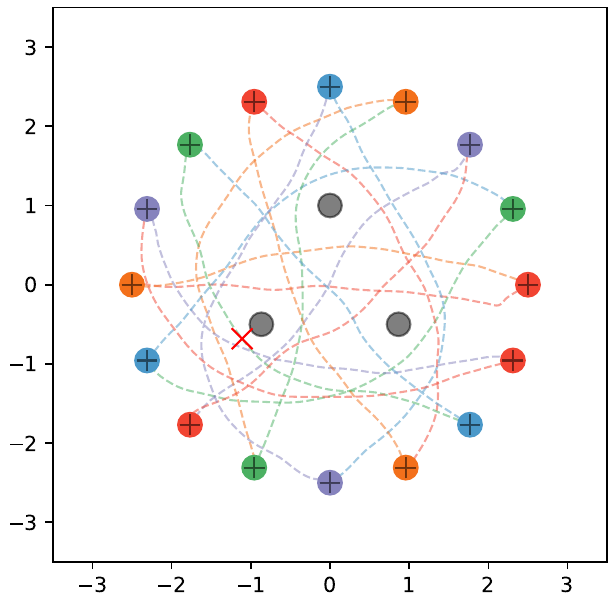}
        {(c) Optimized Coarse Path}%
  \entry{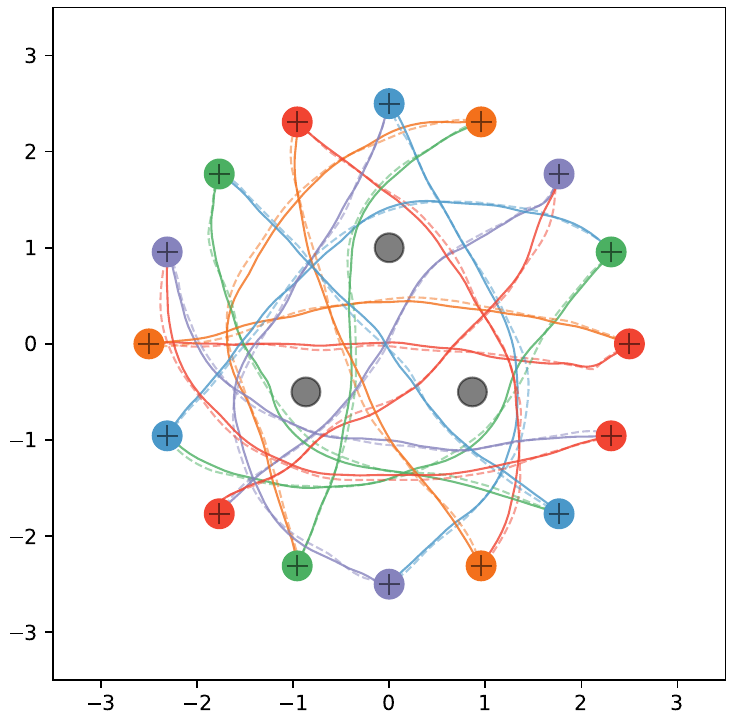}
        {(d) Executed Trajectories}%
  \begin{minipage}[b]{0.31\linewidth}
    \centering
    \begin{tikzpicture}
    \node[anchor=north west] at (0, 0) 
    {\includegraphics[width=\linewidth]{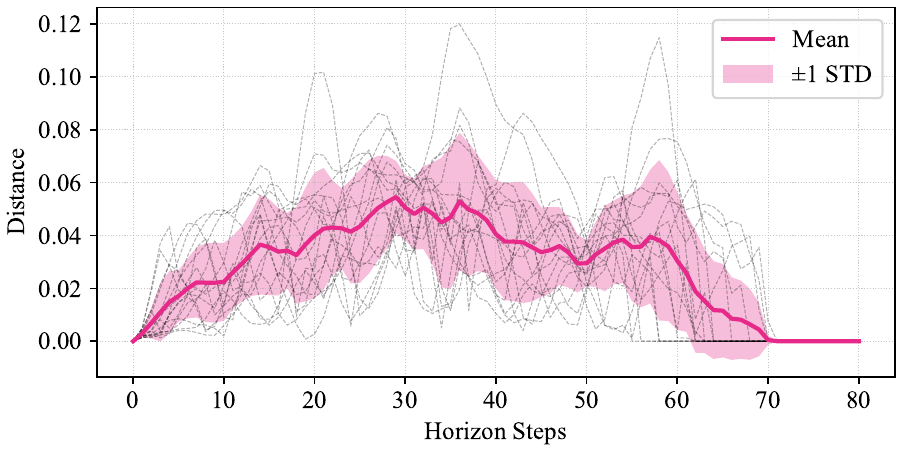}};
    \node[rotate=90] at (0.23, -0.8) {\tiny[\si{\meter}]};
    \end{tikzpicture}
    {\scriptsize (e) Adjustments made in Coarse-to-Fine}
  \end{minipage}
\caption{
Visualization of \textsc{D4orm-C2F}.
The coarse trajectory $\tau\coarse$ has a horizon of $H\coarse$, while the guidance path $p\coarse$ denotes its interpolated form with $H$ waypoints of the positional components.
\emph{(a)} shows the initial guidance $p\coarse$ while \emph{(b)} displays waypoints verified as collision free.
The optimized guidance $p\coarse$ in the backend is presented in (c), which still contains collisions because the optimized coarse trajectory does not guarantee interpolated result $p\coarse$ to be collision-free.
Moreover, it could be kinodynamically infeasible.
\emph{(d)} is the final executed trajectory by following $p\coarse$ in a receding-horizon framework, which eventually resolves collisions.
\emph{(e)} shows the difference between executed trajectory and $p\coarse$, indicating that online planning converts the guidance into a collision-free, executable trajectory with some minor deviations.
}
\label{fig:d4orm_c2f_procedure}
\end{figure*}
}

\subsection{Local Refinement with Receding Horizon}
Once we obtain the guidance path $p\coarse$, we initiate online planning with a receding-horizon scheme (Lines~\ref{algo:d4orm_c2f:online-start}--\ref{algo:d4orm_c2f:online-end}).
At each control iteration, the first step is to identify a joint subgoal $p\subgoal$ within $p\coarse$, determined by averaging the nearest waypoints of each robot and adding a small constant offset $m$ to ensure that all robots progress toward their destinations (\cref{algo:d4orm_c2f:subgoal}).
The subgoals are synchronized across robots in $p\coarse$, thereby preserving the spatiotemporal coordination implied by the coarse guidance.

The next step is to generate a joint collision-free trajectory $\tau\predict$ of horizon $H\predict$ with a time step of $\Delta t$, where $H\predict \ll H$.
We invoke \textsc{D4orm(-D)} with the original reward structure in \cref{eq:generic-reward}, but toward the subgoal $p\subgoal$ (\cref{algo:d4orm_c2f:d4orm}).
The predicted trajectory $\tau\predict$ is then executed for $H\execute$ steps, after which the next control loop begins (\cref{algo:d4orm_c2f:execute}).
Note that $p\subgoal$ serves only as a soft constraint, as these subgoals are not guaranteed to be collision-free.
Instead, the responsibility for collision avoidance is assigned to the fine trajectory generation.

Notably, \textsc{D4orm} used in this phase can exploit trajectories optimized in the previous iteration using a sliding window.
As it synthesizes \emph{deformations}, bootstrapping from the previous solution helps achieve faster convergence towards optimal trajectories while incorporating up-to-date state information.

\subsection{Backend Global Refinement}
While the coarse-to-fine approach helps minimize computational overhead, it is sensitive to the quality of the coarse trajectory $\tau\coarse$ and its consequence $p\coarse$ to extract subgoals.
In particular, if $p\coarse$ incurs a large travel time, following it can lead to redundant motions in the executed trajectory.
Therefore, we introduce a global refinement step that refines the coarse trajectory using \textsc{D4orm}, executed concurrently in the backend alongside online planning (\cref{algo:d4orm_c2f:refine}).
As robots approach their goals, the coarse trajectory optimization becomes progressively easier as remaining planning horizon shortens.

\section{\textsc{D-D4orm}: Distributed Computation}
\label{sec:d4orm_dist}
Although \textsc{D4orm-D} and \textsc{D4orm-C2F} provide viable options for solving large-scale multi-robot trajectory planning, they share an inherent architectural limitation that all computation is performed on a single, centralized computational entity.
This dependency on centralized computation can be prohibitive in resource-constrained environments.
In particular, when the centralized system is not equipped with a highly-capable GPU, it becomes impractical to sample massive numbers of trajectories in real-time to approximate the reward gradient accurately.
Luckily, we recognize that \textsc{D4orm}'s sampling process can be easily \emph{distributed} across computational ``nodes'' (in the sense of a ``cluster'' of computing devices).
This would allow each node to execute \textsc{D4orm} with a smaller sample size, which can potentially even sidestep the need for GPUs.
We therefore develop a distributed version, \textsc{D-D4orm}, in which multiple computing nodes independently perform \textsc{D4orm} while exchanging information through broadcast communication over a fully connected network for all nodes.
The distribution mechanism is designed to evenly split the workload, and can, in principle, accommodate any number of available nodes.
Note that, while ideally each robot acts as a ``node,'' this is not a requirement.

{
\begin{algorithm}[t!]
\caption{\textsc{D-D4orm}: Distributed Planning}
\label{algo:d4orm_dist}
\begin{algorithmic}[1]
\small
\item[{\small\emph{Each node independently runs:}}]
\State Initialize $\tau$
\While{not interrupted}
\State Refine $\tau$ with \textsc{D4orm}
\label{algo:d4orm_dist:refine}
\State Broadcast $\tau$, collect trajectories from others
\State $\Gamma\gets$ the top-$k$ reward trajectories received
\label{algo:d4orm_dist:topk}
\State $\tau \gets$ pickup  one from $\Gamma$ randomly
\label{algo:d4orm_dist:pick}
\EndWhile
\State \Return the best trajectory collected so far
\end{algorithmic}
\end{algorithm}
}

\cref{algo:d4orm_dist} outlines the procedure of \textsc{D-D4orm}.
After initializing a solution trajectory for all robots, each node independently runs \textsc{D4orm} with a different random seed.
The resulting trajectories are then broadcast and collected by all nodes.
Each node then selects the top-$k$ trajectories according to their reward values, randomly chooses one of them as its base trajectory, and runs its next iteration of \textsc{D4orm}.
This process is repeated for a specified number of iterations, or until some user-specified terminal conditions are met.
While \cref{algo:d4orm_dist} uses \textsc{D4orm} (\cref{algo:d4orm_dist:refine}), it is straightforward to use its improved variants such as \textsc{D4orm-D} and \textsc{D4orm-C2F}.

We adopt a top-$k$ reward update (\cref{algo:d4orm_dist:topk}), which favors candidate solutions that have a high reward (i.e., promising) and yet maintain some diversity by picking randomly from this set (\cref{algo:d4orm_dist:pick}).
In \cref{sec:evaluation:distributed}, we empirically compared this approach against two alternatives: the max-reward update ($k{=}1$ in top-$k$), which improves exploitation at the cost of diversity, and the $k$-means update, which improves diversity at the cost of exploitation.

\textit{Distributed on-robot planning:}
As a notable application of \textsc{D-D4orm}, we extend it to the case where every robot in the team acts as a distributed computational node.
Such an onboard implementation is natural, since robots are typically equipped with computing devices, and those may be sufficiently powerful when the problem size shrinks by virtue of being distributed.
Each robot then performs the rollout operation independently on its own device, and subsequently exchanges information with others only after at least one iteration of its denoising process.
This is a practical design choice that avoids excessive frequent inter-robot communication.

\section{Evaluation of \textsc{D4orm} Variants}
\label{sec:evaluation2}
\newcommand{\envlabel}[3][0pt]{%
  \raisebox{#1}{%
    \makebox[0.158\textwidth][c]{%
      \hspace{#2}\small #3%
    }%
  }%
}

\newcommand{\rowlabel}[2][0pt]{%
  \raisebox{#1}{%
    \makebox[2.75mm][r]{%
      \rotatebox{90}{\small #2}%
    }%
  }%
}

\newcommand{\imgwithoverlay}[2]{%
\begin{tikzpicture}[baseline=(img.south)]
  \node[
    anchor=south west,
    inner sep=0pt
  ] (img) at (0,0)
  {\includegraphics[width=0.158\textwidth]{#1}};

  \node[
    anchor=north west,
    inner sep=0pt
  ] at ([xshift=7.0pt,yshift=-0.5pt]img.north west)
  {\includegraphics[width=1.25cm]{#2}};
\end{tikzpicture}%
}

\begin{figure*}[t]
\centering

\setlength{\tabcolsep}{2pt}
\renewcommand{\arraystretch}{1.0}

\resizebox{0.995\textwidth}{!}{%
\begin{tabular}{@{}c *{6}{c}@{}}

&
\envlabel[-1.0mm]{2mm}{Differential Drive} &
\envlabel[-1.0mm]{2mm}{2D Holonomic} &
\envlabel[-1.0mm]{2mm}{Heterogeneous} &
\envlabel[-1.0mm]{2mm}{3D Holonomic} &
\envlabel[-1.0mm]{2mm}{2D Holo 4 Obs} &
\envlabel[-1.0mm]{2mm}{2D Holo Random}
\\[1mm]

\rowlabel[2mm]{$\leftarrow$ runtime [\si{\second}]} &

\includegraphics[width=0.158\textwidth]
{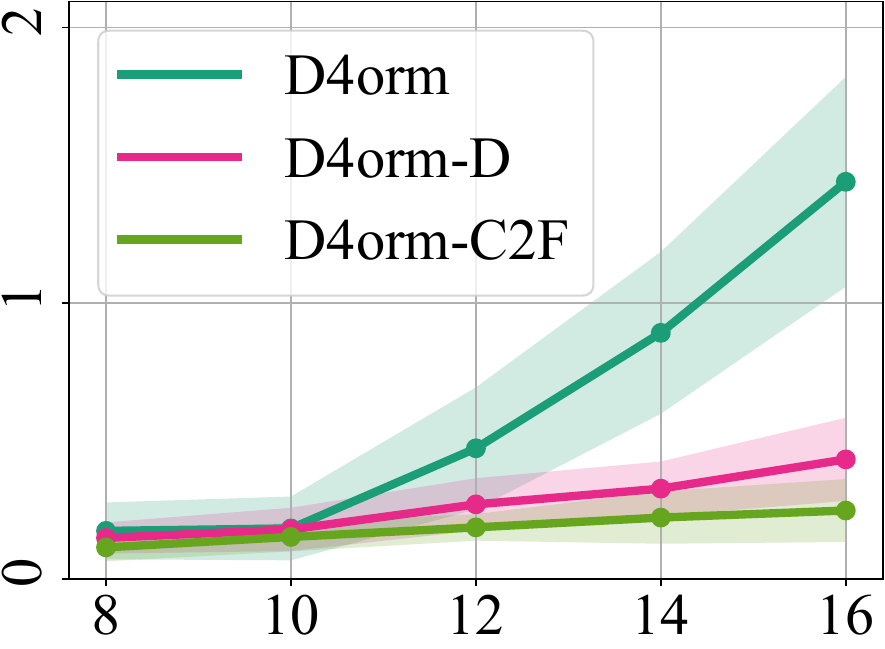}
&

\includegraphics[width=0.158\textwidth]
{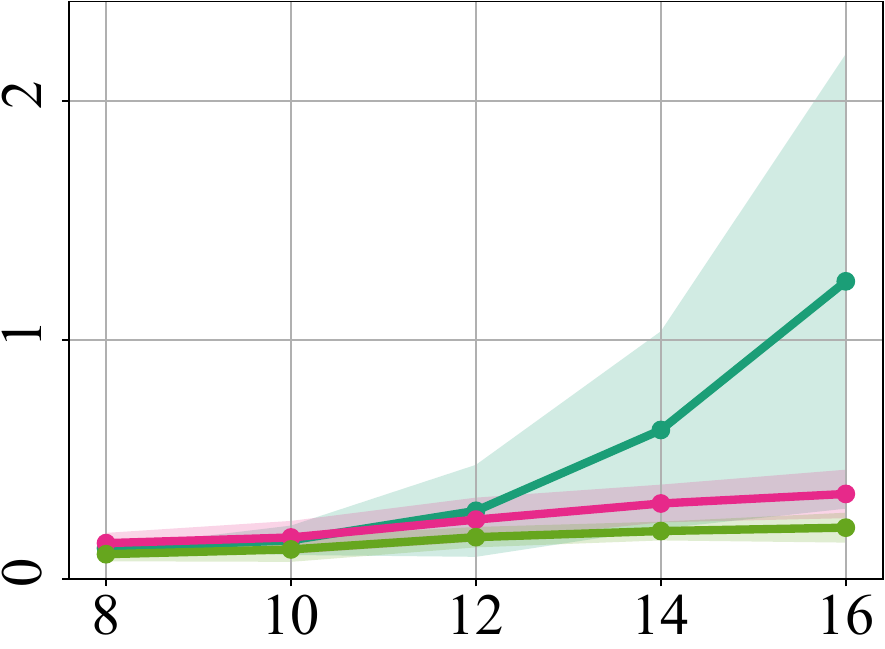}
&

\includegraphics[width=0.158\textwidth]
{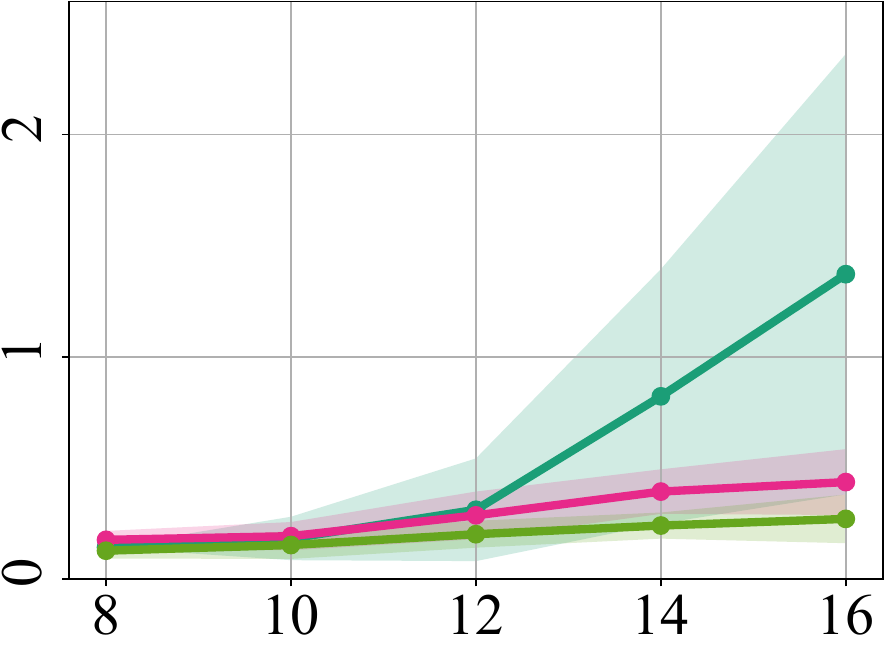}
&

\includegraphics[width=0.158\textwidth]
{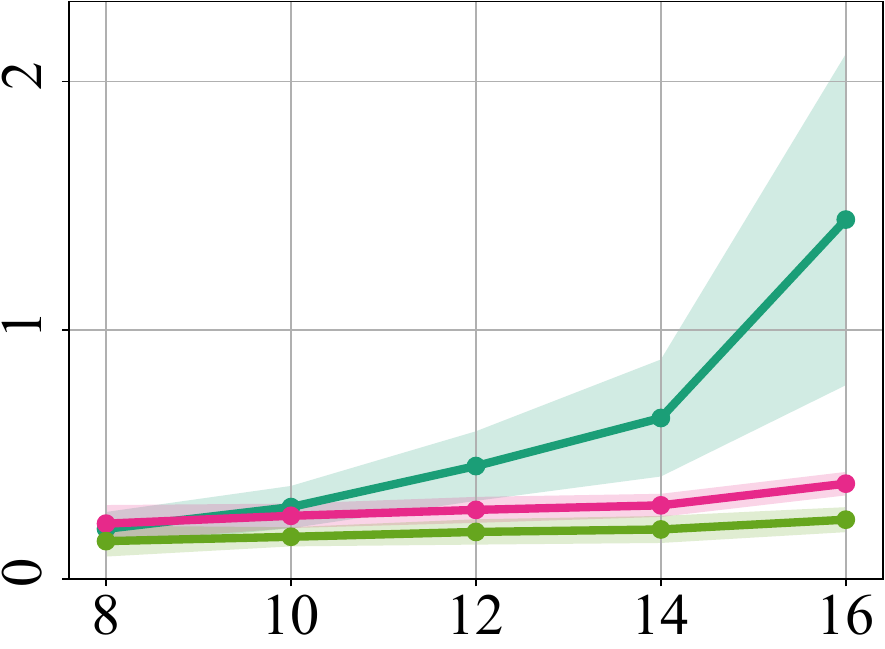}
&

\imgwithoverlay
{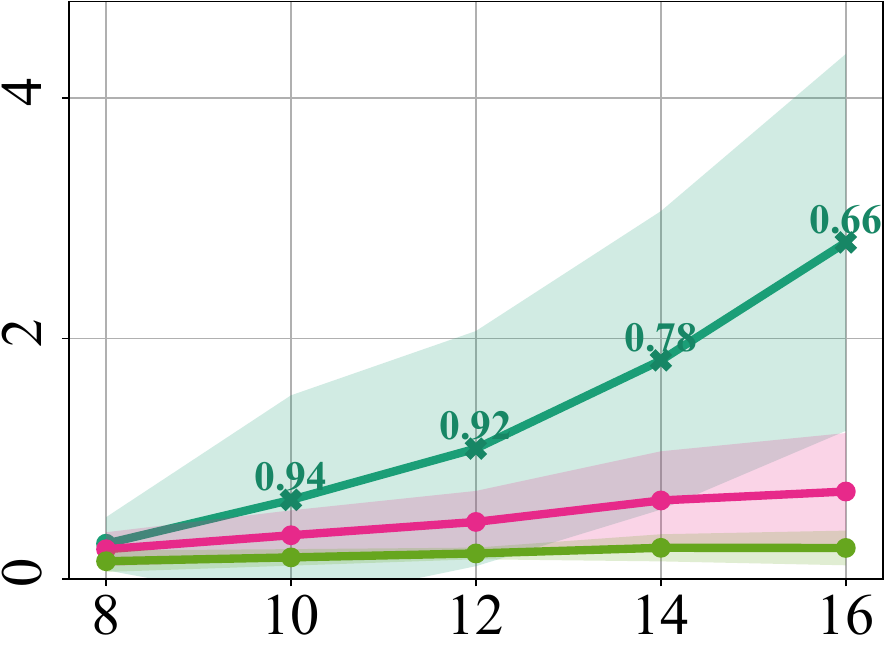}
{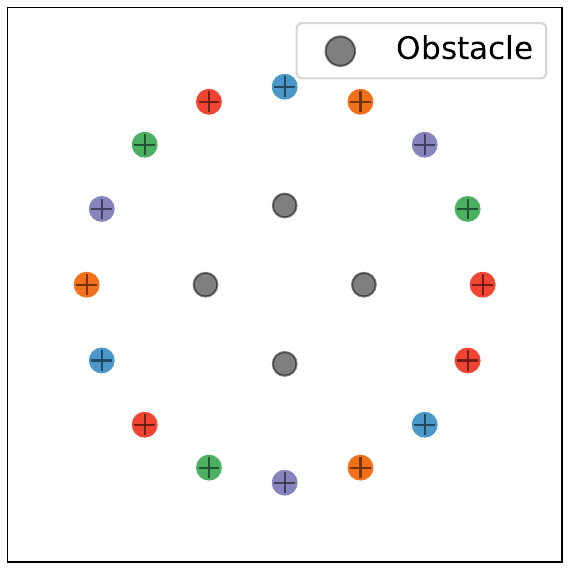}
&

\imgwithoverlay
{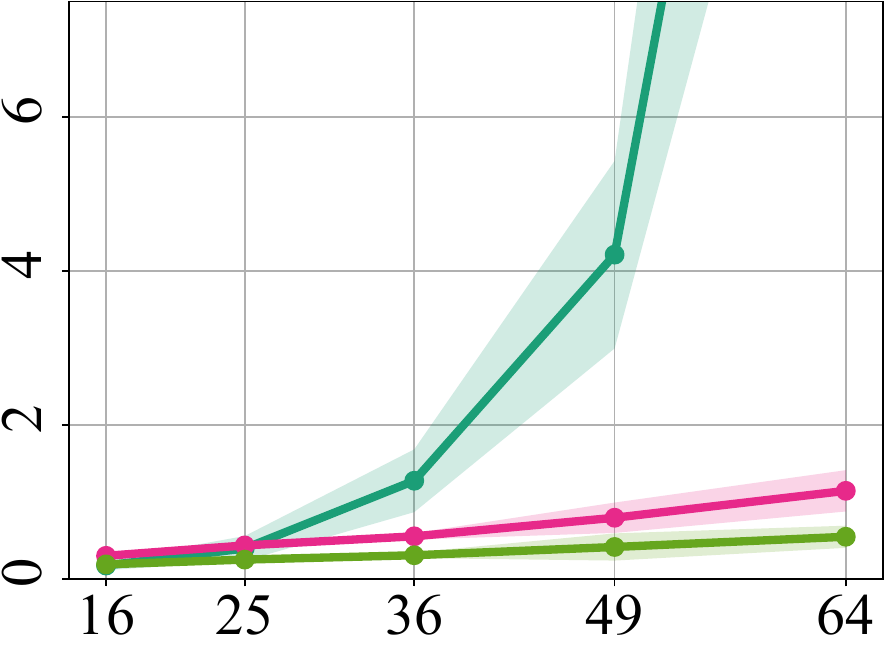}
{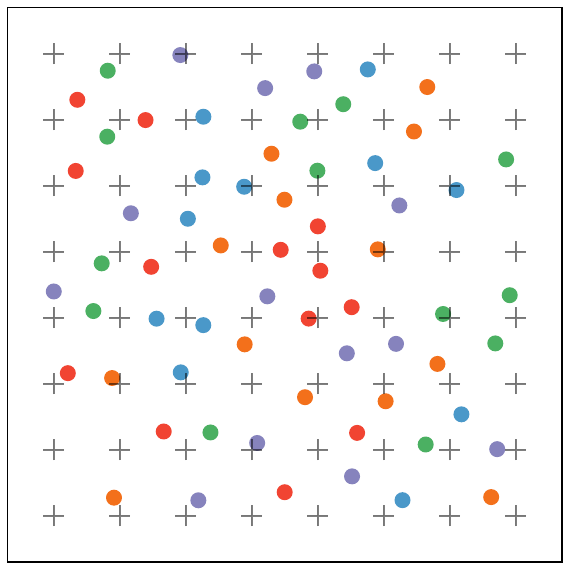}

\\[-1mm]

\rowlabel[0mm]{$\leftarrow$ travel time [s]} &

\includegraphics[width=0.158\textwidth]
{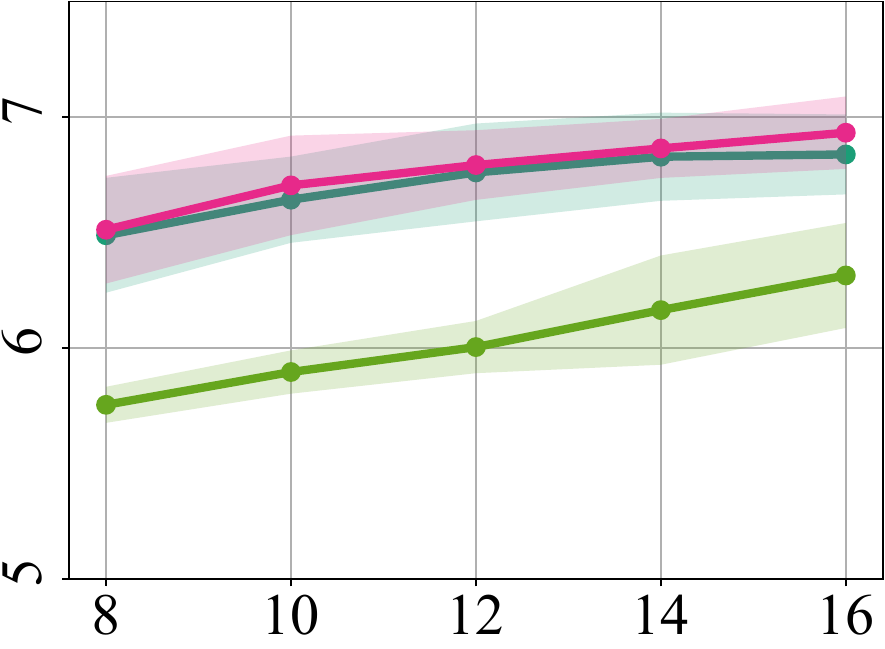}
&

\includegraphics[width=0.158\textwidth]
{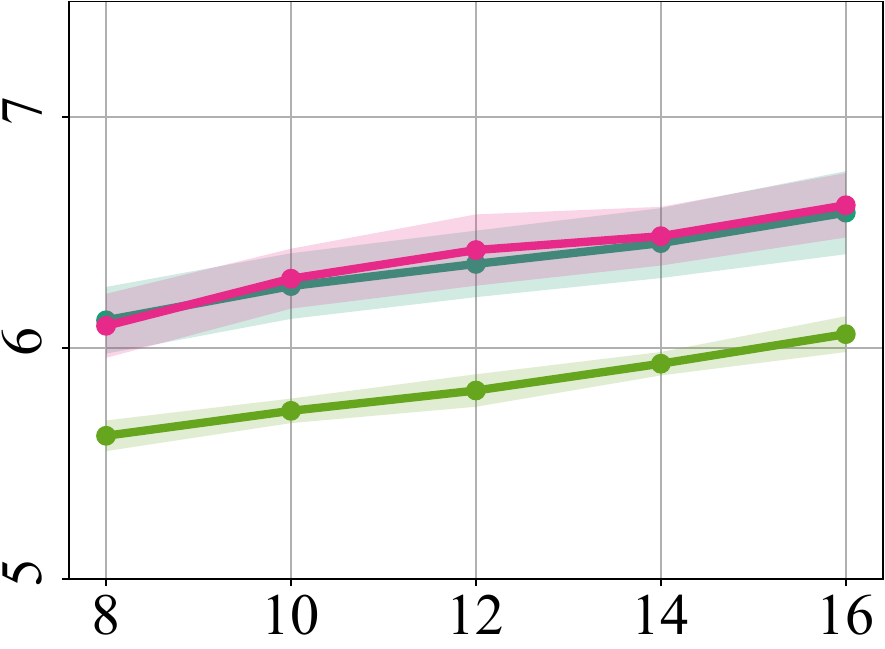}
&

\includegraphics[width=0.158\textwidth]
{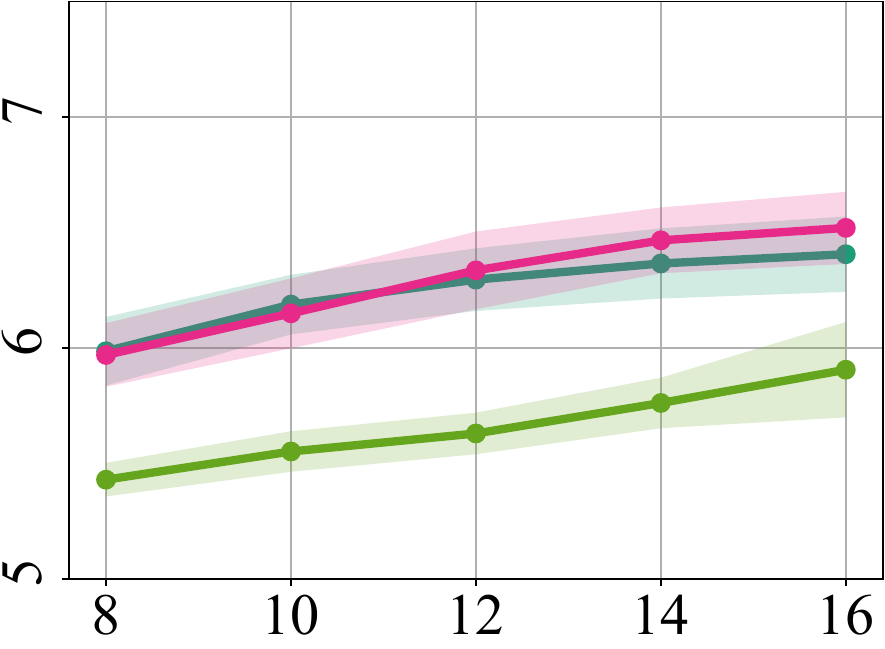}
&

\includegraphics[width=0.158\textwidth]
{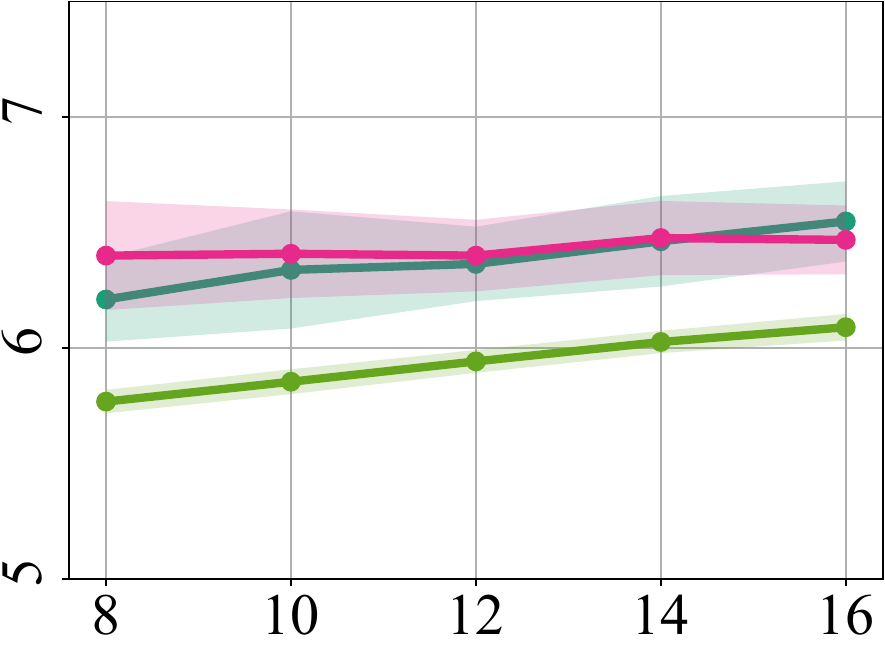}
&

\includegraphics[width=0.158\textwidth]
{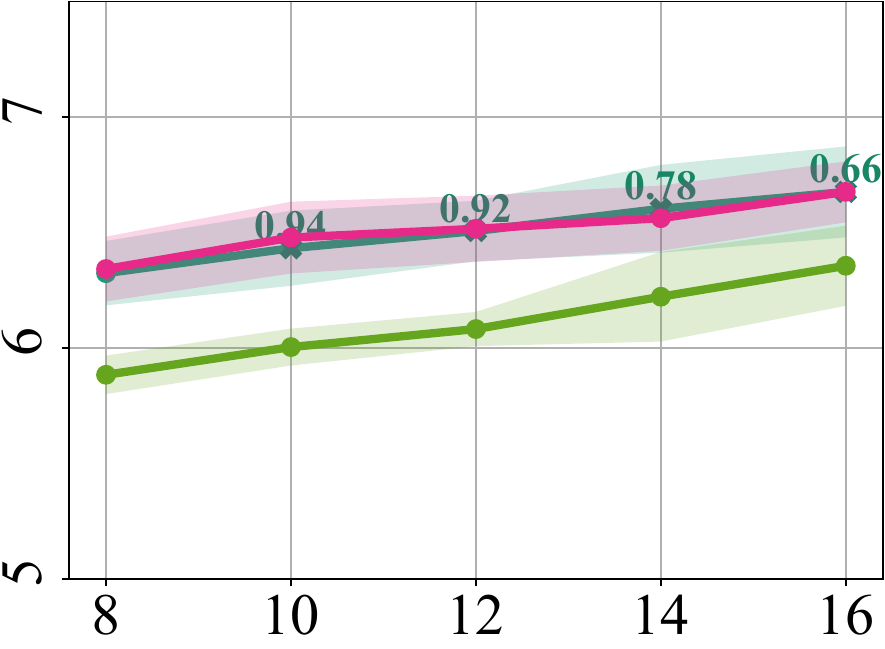}
&

\includegraphics[width=0.158\textwidth]
{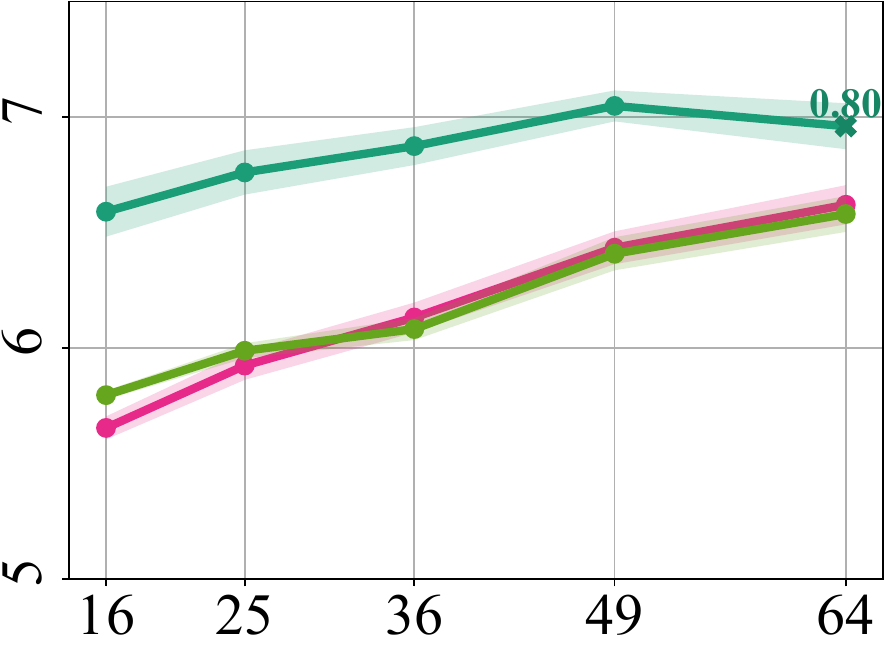}

\\[-1mm]

& \multicolumn{6}{c}{\small number of robots ($\times$: scores obtained by excluding failed runs, accompanied by success rates, $\bullet$: all runs succeeded)}

\end{tabular}%
}

\caption{Comparison of \textsc{D4orm}, \textsc{D4orm-D}, and \textsc{D4orm-C2F} across different environments over 50 trials, using 1,024 samples per step. The number indicates the success rate for configurations that include failures. Runtime for \textsc{D4orm-C2F} is the time required to compute the coarse trajectory. Travel time is the average time required for each robot in the team to reach the goal.}

\label{fig:variants_cmp}

\end{figure*}

We now evaluate variants of \textsc{D4orm} (\textsc{D4orm-D}, \textsc {D4orm-C2F}, \textsc{D-D4orm}), and demonstrate that it serves as a promising tool for solving various trajectory optimization problems, potentially involving large robot teams.
\cref{sec:amswarm} further offers an empirical reference that juxtaposes \textsc{D4orm-D} and AMSwarm~\cite{adajania2023amswarm}, a representative decoupled multi-robot planning method based on numerical optimization.

\subsection{Comparing \textsc{D4orm}, \textsc{D4orm-D}, and \textsc{D4orm-C2F}}
\cref{fig:variants_cmp} presents the empirical performance of \textsc{D4orm} and its variants under the same experimental setup as in \cref{sec:d4orm-analysis:eval}.
We include challenging scenarios for \textsc{D4orm} such as \textbf{2D Holonomic 4 Obstacles}, where \textsc{D4orm} often fails to derive feasible solutions with 16 robots, and \textbf{2D Holonomic Random}, which assigns goal locations randomly, with up to 64 robots.
The runtime report of \textsc{D4orm-C2F} includes both offline and online planning stages.

Overall, both \textsc{D4orm-D} and \textsc{D4orm-C2F} enhance the real-time planning capability compared to the baseline \textsc{D4orm}.
Notably, both extensions reliably generate full-horizon, collision-free trajectories for the 64-robot case within \SI{2}{\second}, despite the large-scale planning problem involving more than $10^4$ optimization variables.

\textsc{D4orm-D} achieves faster planning while maintaining a travel time (i.e., the average time for each robot to reach its goal) comparable to that of \textsc{D4orm}. In the 2D holonomic random scenario with a large number of robots, where coordination is less tightly coupled than in the circular setup, decoupling significantly reduces travel time by effectively alleviating the gradient dilution problem. In addition, the failure-aware replanning mechanism enables \textsc{D4orm-D} to consistently recover feasible solutions, particularly in the four-obstacle scenario.
Its strong planning capability is further evidenced in \cref{fig:overview}(e), where it resolves conflicts for 100~robots within \SI{30}{\second}, demonstrating scalability to very large teams.

\textsc{D4orm-C2F} further accelerates planning speed through the combination of coarse guidance generation and receding-horizon online planning.
Moreover, it achieves significantly lower travel time compared to the other variants.
This can be attributed to two factors:
\emph{(i)}~preparing the guidance path in a decoupled manner and refining it under a coupled representation at the backend during execution, and
\emph{(ii)}~online planning with a short horizon, which allows \textsc{D4orm} to optimize the executing trajectories more effectively.
Typically, offline planning for 16 robots in the 2D holonomic case requires $\approx$\SI{0.2}{\second}, while online planning runs in \SI{0.03}{\second} per control loop, which is sufficiently fast to enable feedback control.

\subsection{Robustness against Perturbation}
\input{figures/d4orm_c2f_noise}
As \textsc{D4orm-C2F} is an online planning method that continuously incorporates state feedback, its execution is less sensitive to tracking errors during execution.
\cref{fig:d4orm_c2f_noise} compares its execution success rate under perturbations introduced as random Gaussian noise applied to the executed controls.
Whereas one-shot offline planning with \textsc{D4orm-D} often fails under such disturbances, \textsc{D4orm-C2F} successfully adapts by replanning at each control iteration, maintaining collision-free and goal-directed behavior.
This property provides the empirical foundation for the onboard distributed execution, which we will demonstrate later.

{

%
\newcommand{\xEnvStart}{0.13}
\newcommand{\xEnvEnd}{1.00}
\newcommand{\nEnvs}{5}

\pgfmathsetmacro{\xEnvWidth}{(\xEnvEnd - \xEnvStart) / \nEnvs}
\pgfmathsetmacro{\xEnvImageWidth}{0.95 * \xEnvWidth}


\newcommand{\xLabelStart}{0.00}
\newcommand{\xLabelEnd}{\xEnvStart}

\pgfmathsetmacro{\xLabelWidth}{\xLabelEnd - \xLabelStart}

\pgfmathsetmacro{\xMethodLabel}{\xLabelStart + 0.95 * \xLabelWidth}
\pgfmathsetmacro{\xMetricLabel}{\xLabelStart + 0.95 * \xLabelWidth}


\newcommand{\yEnvLabel}{3.6}

\newcommand{\yAmsImageTop}{3.4}
\newcommand{\yAmsSuccess}{3.3}
\newcommand{\yAmsMethodLabel}{1.85}
\newcommand{\yAmsRuntime}{-0.2}
\newcommand{\yAmsMissionTime}{-0.6}
\newcommand{\yAmsSmoothness}{-1.0}

\newcommand{\yDPlainImageTop}{-1.4}
\newcommand{\yDPlainSuccess}{-1.5}
\newcommand{\yDPlainMethodLabel}{-3.0}
\newcommand{\yDPlainRuntime}{-5.0}
\newcommand{\yDPlainMissionTime}{-5.4}
\newcommand{\yDPlainSmoothness}{-5.8}

\newcommand{\yDImageTop}{-6.2}
\newcommand{\yDSuccess}{-6.3}
\newcommand{\yDMethodLabel}{-7.8}
\newcommand{\yDRuntime}{-9.8}
\newcommand{\yDMissionTime}{-10.2}
\newcommand{\yDSmoothness}{-10.6}

\newcommand{\yFigureTop}{3.85}
\newcommand{\yFigureBottom}{-10.9}

\newcommand{\na}{\textcolor{gray}{N/A}}


\newcommand{\setEnvX}[1]{%
  \pgfmathsetmacro{\xEnvLeft}{\xEnvStart + (#1 - 1) * \xEnvWidth}%
  \pgfmathsetmacro{\xEnvCenter}{\xEnvLeft + 0.5 * \xEnvWidth}%
  \pgfmathsetmacro{\xSuccess}{\xEnvLeft + 0.06 * \xEnvWidth}%
}

\newcommand{\metricLabels}[5]{%
  \node[anchor=east] at (\xMethodLabel, #1)
    {\normalsize \textbf{#2}};
  \node[anchor=east] at (\xMetricLabel, #3)
    {runtime [\SI{}{\second}]~$\downarrow$};
  \node[anchor=east] at (\xMetricLabel, #4)
    {makespan [\SI{}{\second}]~$\downarrow$};
  \node[anchor=east] at (\xMetricLabel, #5)
    {jerk~$\downarrow$};
}

\newcommand{\successMark}[3]{%
  \ifthenelse{\equal{#3}{1}}{%
    \node[anchor=north west] at (#1, #2)
      {\large \textcolor{teal}{$\checkmark$}};
  }{%
    \node[anchor=north west] at (#1, #2)
      {\large \textcolor{red}{$\times$}};
  }%
}

\newcommand{\envLabel}[1]{%
  \setEnvX{#1}
  \node at (\xEnvCenter, \yEnvLabel) {env\_#1};
}

\newcommand{\envLabelRow}{%
  \envLabel{1}
  \envLabel{2}
  \envLabel{3}
  \envLabel{4}
  \envLabel{5}
}

\newcommand{\methodMetrics}[8]{%
  \successMark{\xSuccess}{#1}{#5}
  \node at (\xEnvCenter, #2) {#6};
  \node at (\xEnvCenter, #3) {#7};
  \node at (\xEnvCenter, #4) {#8};
}

\newcommand{\methodCell}[8]{%
  \setEnvX{#1}

  \node[anchor=north, inner sep=0pt] at (\xEnvCenter, #3)
    {\includegraphics[
      width=\xEnvImageWidth\linewidth,
      trim={0 0 0 0},
      clip
    ]{images/ams_d4orm_cmp/#2_#1}};

  \methodMetrics{#4}{#5}{#6}{#7}#8
}

\newcommand{\methodRowAMSwarm}[5]{%
  \metricLabels{\yDMethodLabel}
               {AMSwarm}
               {\yDRuntime}
               {\yDMissionTime}
               {\yDSmoothness}
  \methodCell{1}{ams}{\yDImageTop}{\yDSuccess}{\yDRuntime}{\yDMissionTime}{\yDSmoothness}#1
  \methodCell{2}{ams}{\yDImageTop}{\yDSuccess}{\yDRuntime}{\yDMissionTime}{\yDSmoothness}#2
  \methodCell{3}{ams}{\yDImageTop}{\yDSuccess}{\yDRuntime}{\yDMissionTime}{\yDSmoothness}#3
  \methodCell{4}{ams}{\yDImageTop}{\yDSuccess}{\yDRuntime}{\yDMissionTime}{\yDSmoothness}#4
  \methodCell{5}{ams}{\yDImageTop}{\yDSuccess}{\yDRuntime}{\yDMissionTime}{\yDSmoothness}#5
}

\newcommand{\methodRowDFourorm}[5]{%
  \metricLabels{\yAmsMethodLabel}
               {\textsc{D4orm}}
               {\yAmsRuntime}
               {\yAmsMissionTime}
               {\yAmsSmoothness}
  \methodCell{1}{d4orm}{\yAmsImageTop}{\yAmsSuccess}{\yAmsRuntime}{\yAmsMissionTime}{\yAmsSmoothness}#1
  \methodCell{2}{d4orm}{\yAmsImageTop}{\yAmsSuccess}{\yAmsRuntime}{\yAmsMissionTime}{\yAmsSmoothness}#2
  \methodCell{3}{d4orm}{\yAmsImageTop}{\yAmsSuccess}{\yAmsRuntime}{\yAmsMissionTime}{\yAmsSmoothness}#3
  \methodCell{4}{d4orm}{\yAmsImageTop}{\yAmsSuccess}{\yAmsRuntime}{\yAmsMissionTime}{\yAmsSmoothness}#4
  \methodCell{5}{d4orm}{\yAmsImageTop}{\yAmsSuccess}{\yAmsRuntime}{\yAmsMissionTime}{\yAmsSmoothness}#5
}

\newcommand{\methodRowDFourormD}[5]{%
  \metricLabels{\yDPlainMethodLabel}
               {\textsc{D4orm-D}}
               {\yDPlainRuntime}
               {\yDPlainMissionTime}
               {\yDPlainSmoothness}
  \methodCell{1}{d4orm_d}{\yDPlainImageTop}{\yDPlainSuccess}{\yDPlainRuntime}{\yDPlainMissionTime}{\yDPlainSmoothness}#1
  \methodCell{2}{d4orm_d}{\yDPlainImageTop}{\yDPlainSuccess}{\yDPlainRuntime}{\yDPlainMissionTime}{\yDPlainSmoothness}#2
  \methodCell{3}{d4orm_d}{\yDPlainImageTop}{\yDPlainSuccess}{\yDPlainRuntime}{\yDPlainMissionTime}{\yDPlainSmoothness}#3
  \methodCell{4}{d4orm_d}{\yDPlainImageTop}{\yDPlainSuccess}{\yDPlainRuntime}{\yDPlainMissionTime}{\yDPlainSmoothness}#4
  \methodCell{5}{d4orm_d}{\yDPlainImageTop}{\yDPlainSuccess}{\yDPlainRuntime}{\yDPlainMissionTime}{\yDPlainSmoothness}#5
}


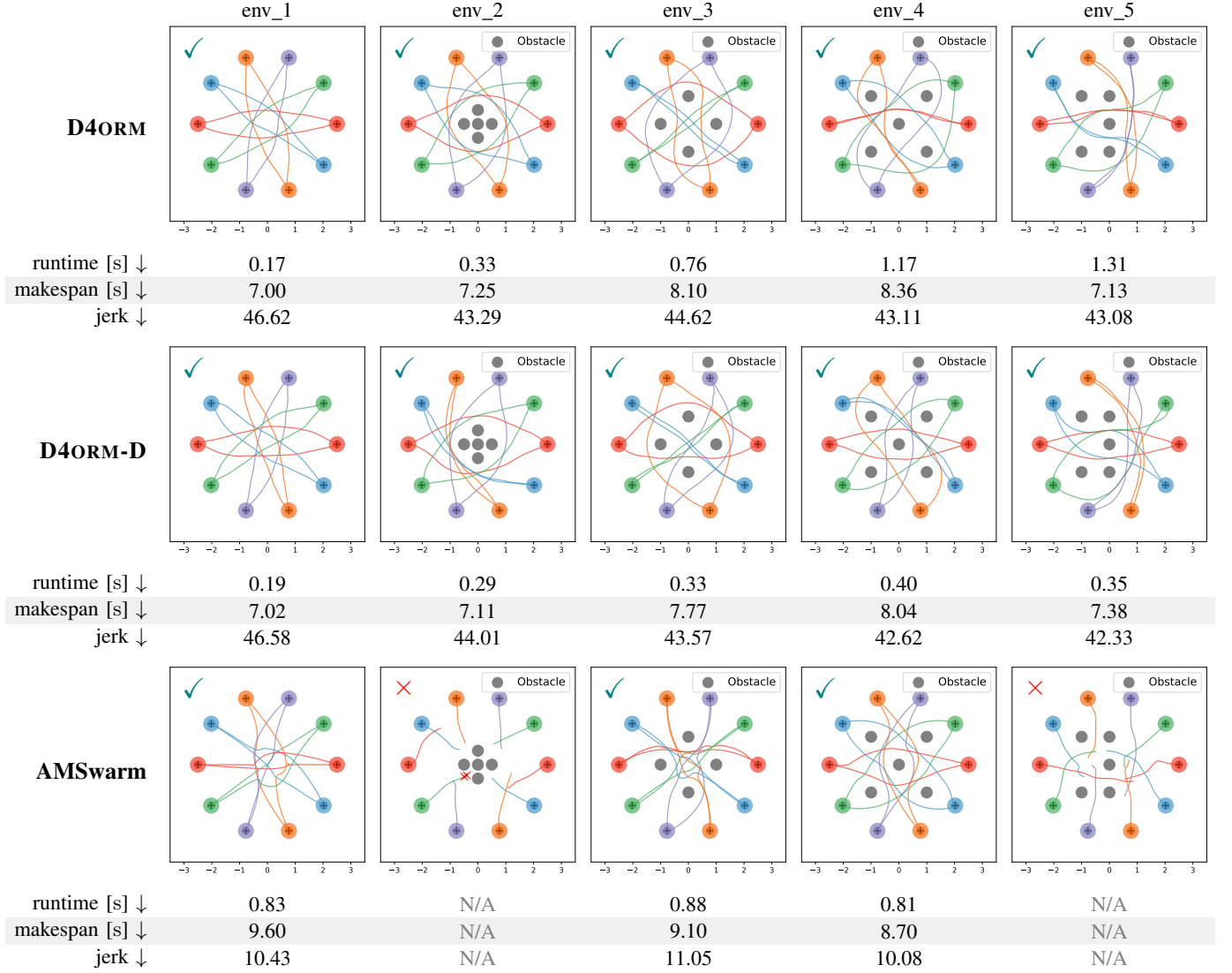
\begin{figure*}[th!]
  \centering

  \begin{tikzpicture}[x=\linewidth, y=1cm]
    \small

    \definecolor{bgcolor}{RGB}{240, 240, 240}

    \path[fill=bgcolor]
      (0.0, \yAmsMissionTime - 0.2)
      rectangle
      (1.0, \yAmsMissionTime + 0.2);

    \path[fill=bgcolor]
      (0.0, \yDPlainMissionTime - 0.2)
      rectangle
      (1.0, \yDPlainMissionTime + 0.2);

    \path[fill=bgcolor]
      (0.0, \yDMissionTime - 0.2)
      rectangle
      (1.0, \yDMissionTime + 0.2);

    \envLabelRow

    \methodRowAMSwarm
      {{1}{0.83}{9.60}{10.43}}
      {{0}{\na}{\na}{\na}}
      {{1}{0.88}{9.10}{11.05}}
      {{1}{0.81}{8.70}{10.08}}
      {{0}{\na}{\na}{\na}}

    \methodRowDFourorm
      {{1}{0.17}{7.00}{46.62}}
      {{1}{0.33}{7.25}{43.29}}
      {{1}{0.76}{8.10}{44.62}}
      {{1}{1.17}{8.36}{43.11}}
      {{1}{1.31}{7.13}{43.08}}

    \methodRowDFourormD
      {{1}{0.19}{7.02}{46.58}}
      {{1}{0.29}{7.11}{44.01}}
      {{1}{0.33}{7.77}{43.57}}
      {{1}{0.40}{8.04}{42.62}}
      {{1}{0.35}{7.38}{42.33}}

    \pgfresetboundingbox
    \path[use as bounding box]
      (0.0, \yFigureBottom) rectangle (1.0, \yFigureTop);

  \end{tikzpicture}

  \caption{
    Juxtaposing trajectories from \textsc{D4orm} and \textsc{D4orm-D}
    against AMSwarm~\cite{adajania2023amswarm}, a representative
    receding-horizon decoupled planning method, using ten 3D holonomic
    robots for five scenarios.
    The plots are top-down 2D projections.
    Symbols \textcolor{teal}{$\checkmark$} and \textcolor{red}{$\times$}
    show whether each method has solved the instance.
    ``jerk'' represents the mean squared jerk.
  }
  \label{fig:vs_amswarm}
\end{figure*}
}
\subsection{Juxtaposing against AMSwarm}
\label{sec:amswarm}
We now present an empirical reference that juxtaposes \textsc{D4orm(-D)} and AMSwarm~\cite{adajania2023amswarm}, a representative multi-robot planning method based on numerical optimization.
AMSwarm differs significantly from \textsc{D4orm-D} in that it adopts robot-wise decoupled, receding-horizon planning, where each robot uses neighbors’ previous trajectories as predictions.
Thus, the comparison is not meant to be fair, and is rather meant to illustrate qualitative differences between these approaches.
We use the publicly available AMSwarm implementation without any customizations, and only adjust parameters to match the configuration adopted by \textsc{D4orm(-D)}.

\cref{fig:vs_amswarm} presents five case studies with ten 3D holonomic robots and cylindrical obstacles (when present). Both \textsc{D4orm} and \textsc{D4orm-D} solve all instances and achieve comparable metrics in easier environments.
In contrast, AMSwarm can be unable to reason about global preferences of other robots and can get stuck in local minima that sometimes lead to failure (see \texttt{env\_2} and \texttt{env\_5}). In general, we see that \textsc{D4orm-D} yields reliable and efficient solutions across \emph{all} test cases with lower computation time and makespan (i.e., mission completion time). This is primarily because of the full-horizon planning performed by \textsc{D4orm-D} and its effective use of GPU acceleration.
We further evaluate trajectory smoothness using mean squared jerk at $\Delta t{=}0.1$. AMSwarm generally performs better on this metric because \textsc{D4orm(-D)} does not include jerk minimization in the objective function. Instead \textsc{D4orm(-D)} prioritizes travel-time reduction and thus produces more direct, less circuitous trajectories. Notably, low jerk values can also be readily achieved in \textsc{D4orm(-D)} by adding jerk limits to the current robot dynamics.

{
\newcommand{\distentry}[2]{%
  \begin{minipage}[b]{0.19\linewidth}
    \centering
    \raisebox{0.1cm}{\hspace*{3mm}\small #2}
    \smallskip\\
    \includegraphics[
      width=\linewidth,
      keepaspectratio
    ]{images/d4orm_dist_cmp/dist_itr_cmp_#1}
  \end{minipage}%
}

\newcommand{\distentryfix}[1]{%
  \begin{minipage}[b]{0.19\linewidth}
    \centering
    \includegraphics[
      width=\linewidth,
      keepaspectratio
    ]{images/d4orm_dist_cmp/top4_itr_cmp_#1}
  \end{minipage}%
}

\newcommand{\solidline}{%
    \raisebox{0.5ex}{\tikz{\draw[line width=1pt] (0,0) -- (0.3,0);}}%
}

\newcommand{\dashedline}{%
    \raisebox{0.5ex}{\tikz{\draw[dashed,line width=1pt] (0,0) -- (0.35,0);}}%
}

\begin{figure*}[t!]
  \centering

  \begin{minipage}{0.99\textwidth}
  \centering

  \begin{minipage}[b]{0.025\linewidth}
    \begin{tikzpicture}
      \small
      \node[rotate=90] at (0, 1.3)
        {$\leftarrow$~iterations};
      \node[] at (0, 0) {};
    \end{tikzpicture}
  \end{minipage}%
  \hspace{0.005\linewidth}%
  \distentry{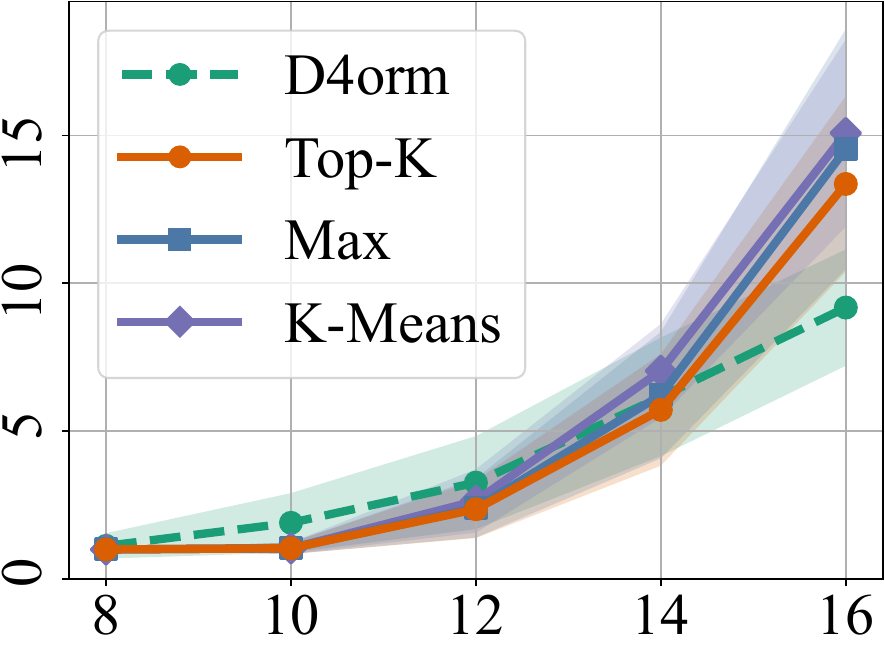}{Differential Drive}%
  \hfill
  \distentry{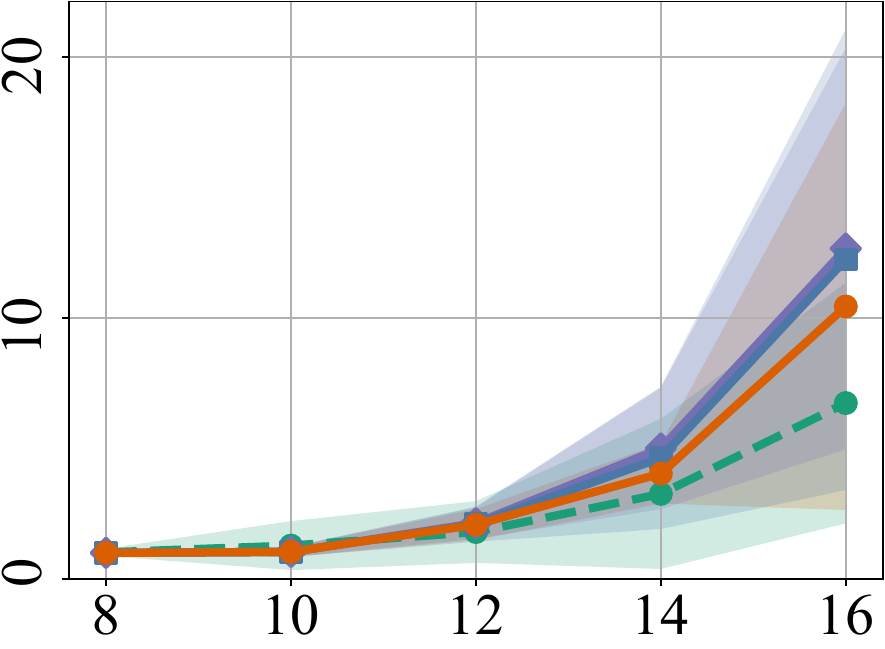}{2D Holonomic}%
  \hfill
  \distentry{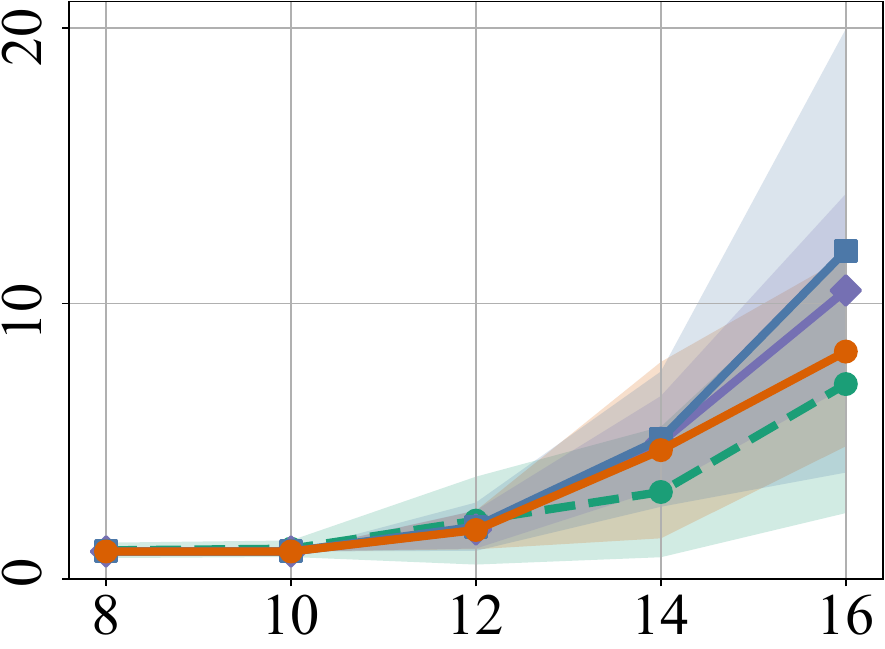}{2D Heterogeneous}%
  \hfill
  \distentry{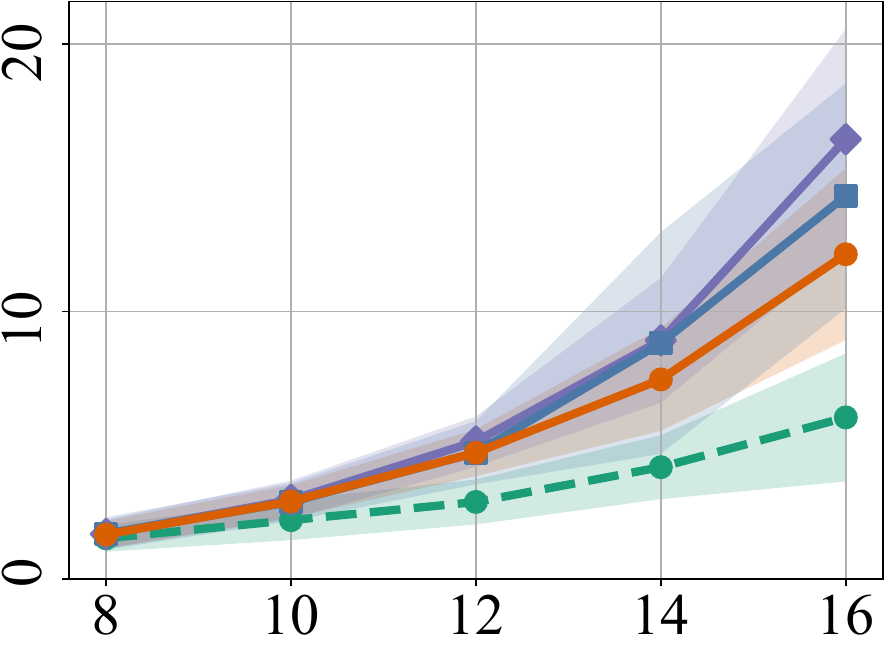}{3D Holonomic}

  \par\vspace{-1mm}

  {\small
  \textbf{\#robots}
  (\dashedline: one centralized computing node, \solidline: \#computing nodes equal to \#robots)
  }

  \par\vspace{2mm}

  \begin{minipage}[b]{0.025\linewidth}
    \begin{tikzpicture}
      \small
      \node[rotate=90] at (0, 1.3)
        {$\leftarrow$~iterations};
      \node[] at (0, 0) {};
    \end{tikzpicture}
  \end{minipage}%
  \hspace{0.005\linewidth}%
  \distentryfix{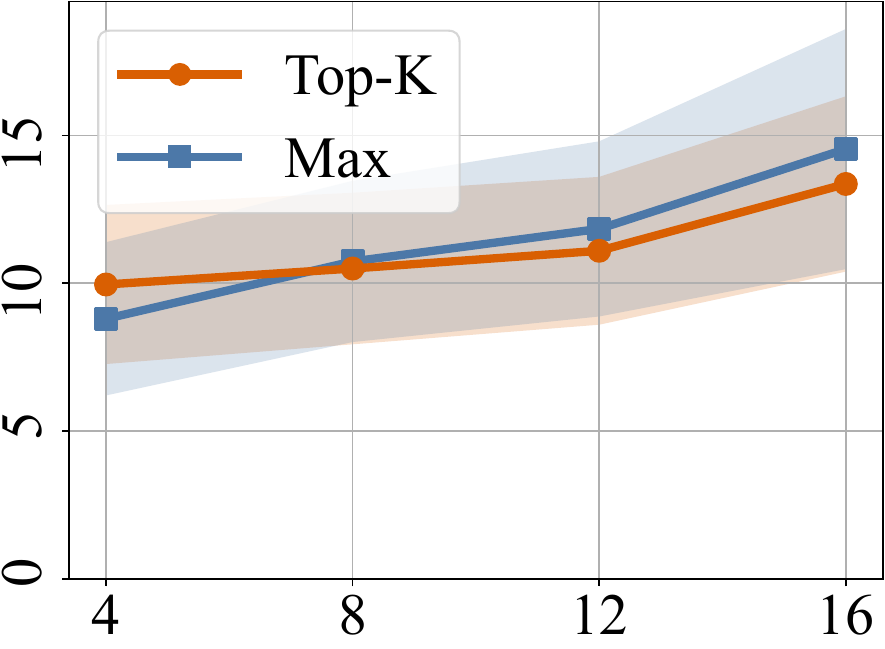}%
  \hfill
  \distentryfix{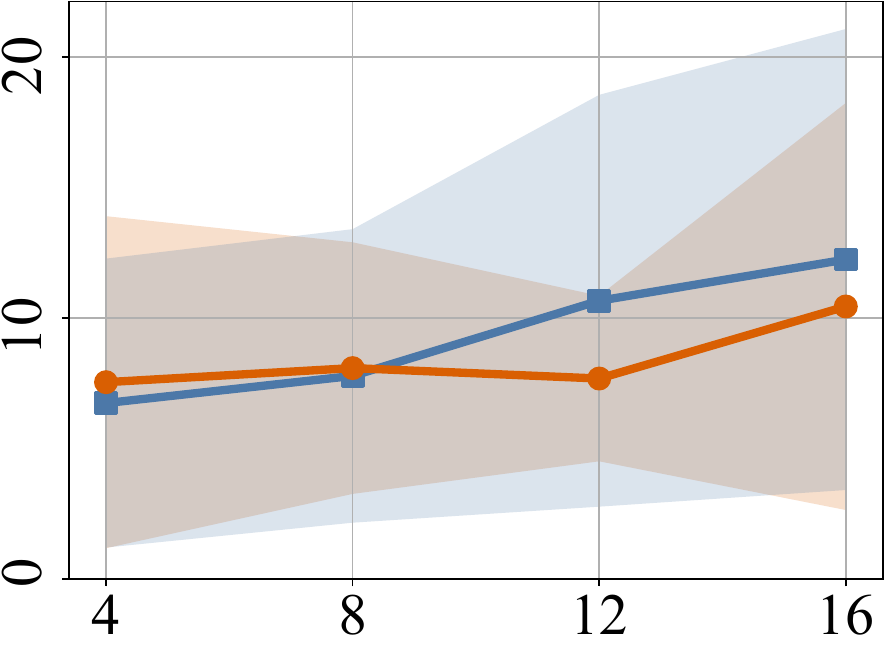}%
  \hfill
  \distentryfix{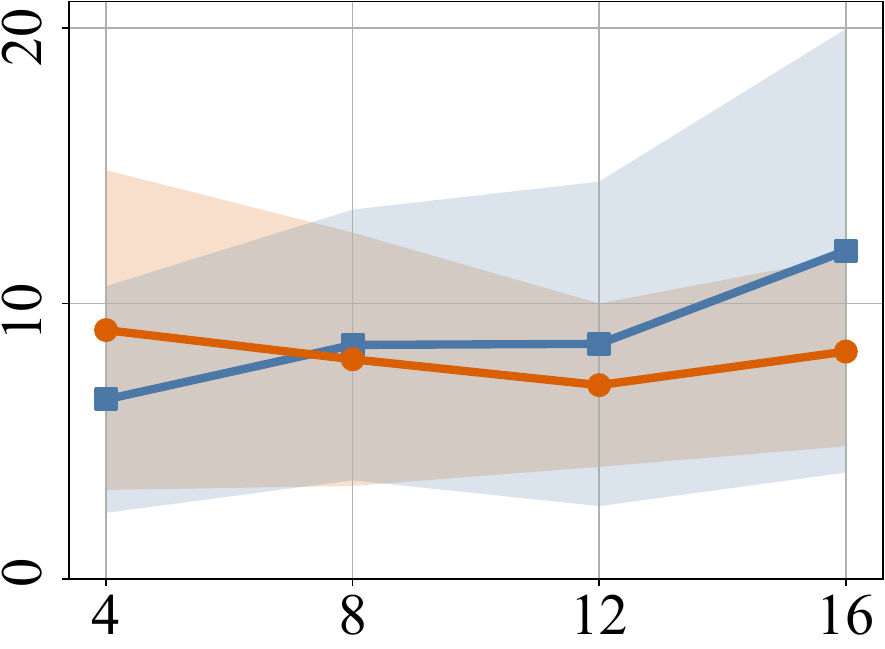}%
  \hfill
  \distentryfix{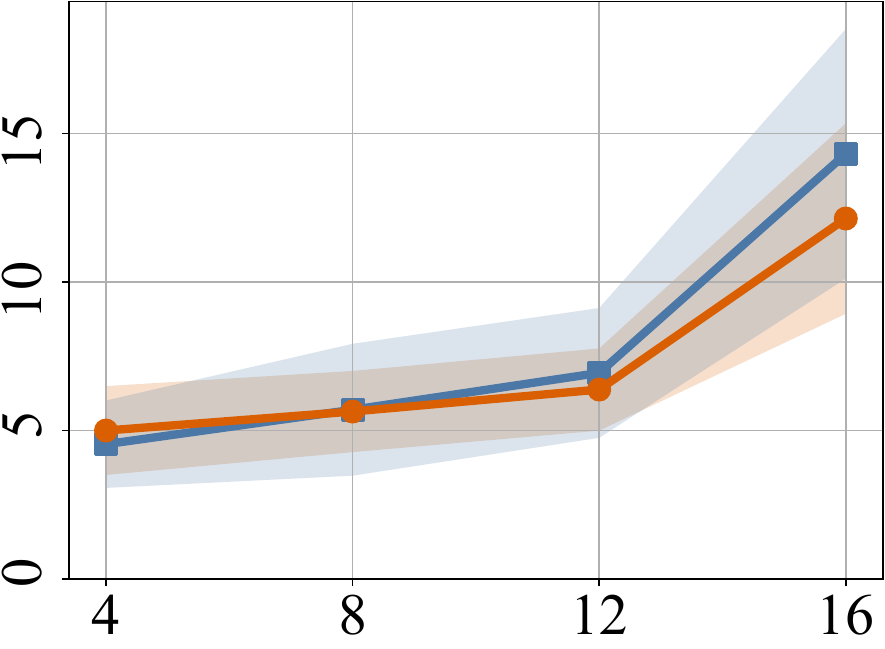}

  \par\vspace{-1mm}

  {\small
  \textbf{\#computing nodes}
  (fix \#robots equal to $16$)
  }

  \end{minipage}

  \caption{
  Evaluation of \textsc{D-D4orm}.
  The number of iterations required to obtain a feasible solution is shown for each environment.
  Each denoising optimization uses a total of 2,048 Monte Carlo samples.
  The top row illustrates the case when each robot is also a computing node, i.e., the size of the problem increases with the number of robots, while the number of samples at each node decreases (e.g., each robot has $M{=}128$ samples in the 16-robot case).
  The bottom row illustrates the effect of varying only the number of computing nodes for a problem that contains a fixed number of robots (${=}16$) (e.g., each node has $M{=}256$ samples in the 8-node case).
  }
  \label{fig:d4orm_dist_cmp}

\end{figure*}
}

\subsection{Distributed Implementation Designs}
\label{sec:evaluation:distributed}
We next investigate the behavior of \textsc{D-D4orm}, the distributed variant of \textsc{D4orm}, using the same total number of samples (${=}2{,}048$).
Recall that in each computing node, \cref{algo:d4orm_dist} employs a top-$k$ update strategy for its next iteration, selecting a solution from the pool of candidate solutions broadcast by other nodes in the previous iteration.
However, as discussed earlier, it is possible to consider other selection strategies.

\cref{fig:d4orm_dist_cmp} (upper) first compares the original \textsc{D4orm} (without distributed computation, with the same total number of samples), \textsc{D-D4orm} with top-$k$ update ($k{=}4$), and two alternative update schemes:
``max,'' which selects only the single best candidate (i.e., top-$1$), and ``$k$-means,'' which selects $k$ diverse candidates based on $k$-means clustering, where the diversity is measured as the $\ell_2$-norm between two candidate solutions.
The results indicate that the top-$k$ strategy, despite its simplicity, consistently achieves the best balance between exploitation and diversity, yielding superior overall performance compared to the other update rules.
While \textsc{D-D4orm} typically requires more iterations than the original \textsc{D4orm} to find a solution under an equivalent total sample budget, distributed computation enables greater flexibility in implementation and allows the total number of samples to scale with the number of participating nodes in practice.

We are further interested in evaluating the impact of changing the number of computing nodes while keeping the number of total Monte Carlo samples constant.
\cref{fig:d4orm_dist_cmp} (lower) shows this effect in the performance of \textsc{D-D4orm} with the top-$k$ ($k{=}4$) and max strategies, by changing the number of nodes and keeping the total number of Monte Carlo samples fixed (${=}2{,}048$).
Intuitively, when the work is split evenly across nodes, every node solves a tiny problem quickly, but potentially at the expense of a slightly inferior solution quality.
We see this evidenced in our evaluations: dividing the samples among a larger number of nodes leads to an increase in the required iterations for both strategies.
The top-$k$ strategy yet remains more effective than the ``max'' when \#nodes is greater than $k$.
We therefore posit that more sophisticated work distribution strategies can be employed if the situation demands, although that is out of the scope of our focus here.

\subsection{Onboard Distributed Demonstration}
{
\newcommand{\ybtm}{3.7}
\begin{figure*}[t]
\centering
\begin{tikzpicture}
  \small
  \node[anchor=south west] at (0, 0)
       {\includegraphics[width=0.95\linewidth]{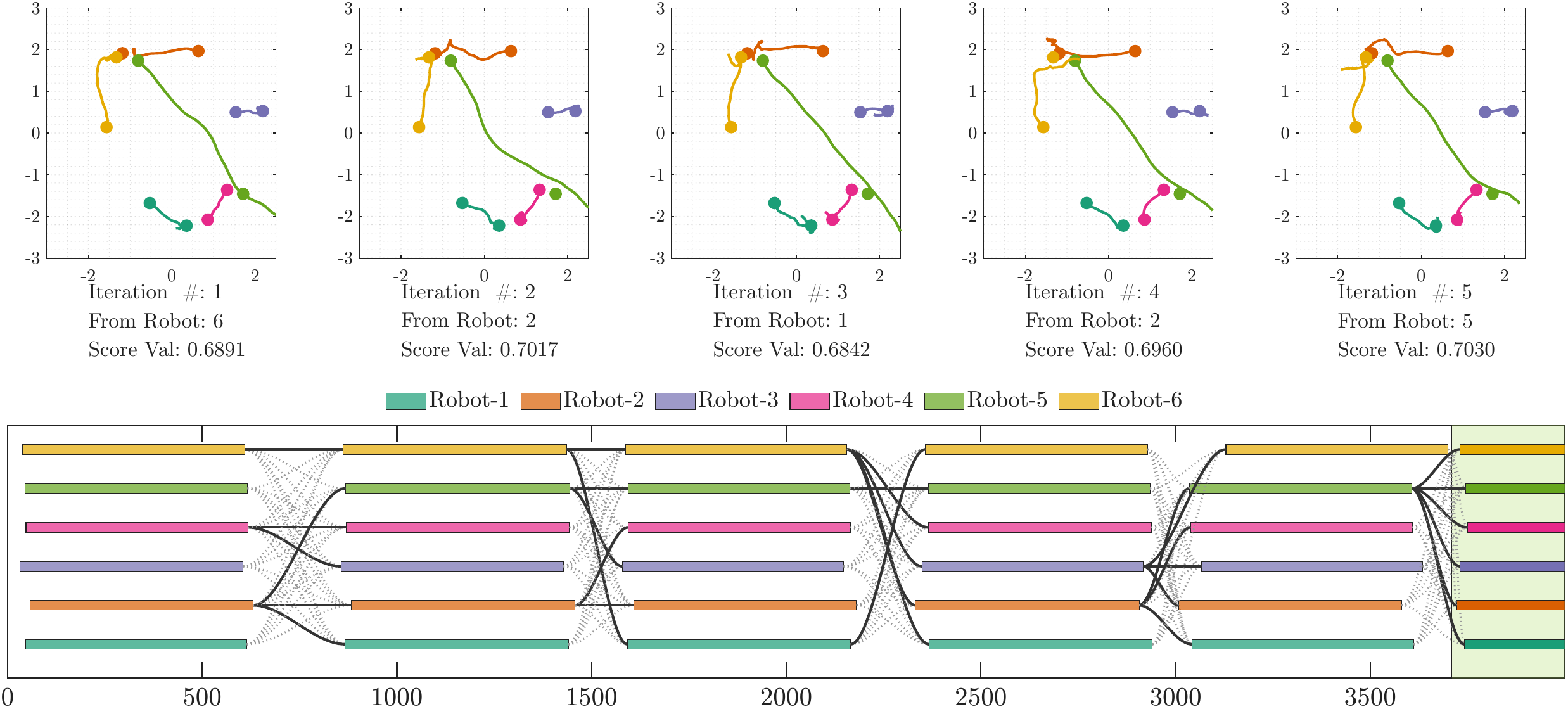}};
  \path[fill=white]
  (0.0, 3.3) --++ (1.0\linewidth, 0) --++ (0, 0.4) --++ (-1.0\linewidth, 0) -- cycle;
  \node[anchor=south] at (0.5\linewidth, -0.4) {time [\SI{}{\milli\second}]};
  \node[anchor=south, rotate=90] at (-0.15, 1.8) {robot};
  \node[anchor=west] at (-0.2, 0.8) {1};
  \node[anchor=west] at (-0.2, 1.25) {2};
  \node[anchor=west] at (-0.2, 1.7) {3};
  \node[anchor=west] at (-0.2, 2.1) {4};
  \node[anchor=west] at (-0.2, 2.55) {5};
  \node[anchor=west] at (-0.2, 3.0) {6};
  \node[anchor=south east] at (0.96\linewidth, 3.2) {execution};

  \path[draw=lightgray,solid,line width=1pt] (2.7, 3.0) -- (1.8, \ybtm);
  \path[draw=lightgray,solid,line width=1pt] (6.4, 1.3) -- (5.4, \ybtm);
  \path[draw=lightgray,solid,line width=1pt] (9.4, 0.8) -- (8.8, \ybtm);
  \path[draw=lightgray,solid,line width=1pt] (12.6, 1.3) -- (12.2, \ybtm);
  \path[draw=lightgray,solid,line width=1pt] (15.6, 2.6) -- (15.5, \ybtm);
\end{tikzpicture}
\caption{
Process visualization for \textsc{D-D4orm}, the distributed onboard planner with six robots {\setlength\fboxsep{0pt}%
\fbox{%
\textcolor[HTML]{1b9e77}{ \rule{7pt}{5pt} }%
\textcolor[HTML]{d95f02}{\rule{7pt}{5pt} }%
\textcolor[HTML]{7570b3}{\rule{7pt}{5pt} }%
\textcolor[HTML]{e7298a}{\rule{7pt}{5pt} }%
\textcolor[HTML]{66a61e}{\rule{7pt}{5pt} }%
\textcolor[HTML]{e6ab02}{\rule{7pt}{5pt} }%
}} that also work as computational nodes, deployed in a real-world lifelong scenario.
Over successive iterations, \textsc{D-D4orm} improves the solution by exchanging candidate solutions computed by each node (i.e., robot), even though a feasible solution is obtained in the first iteration.
A problem is generated at time $t{=}0$, and the bottom row illustrates the planning process for each node and how messages are exchanged over time (light gray).
The solution that a node selected for its next iteration (from top-$k$) is highlighted with black connectors.
The top row visualizes the best intermediate solutions from each iteration.
}
\label{fig:real-world-dist}
\end{figure*}
}

Finally, we evaluate the most notable deployment strategy that employs all robots in a team as computing nodes to run \textsc{D-D4orm}.
We use a team of Cambridge RoboMasters~\cite{blumenkamp2024cambridge}, which are omnidirectional ground robots equipped with mecanum wheels.
Each robot is modeled as a first-order 2D holonomic system, receives fixed-frame positioning data from a motion-capture system over Wi-Fi, and runs its local feedback-control stack onboard its NVIDIA Jetson Orin System-on-Chip.
The entire control and estimation stack is implemented in C++ over ROS~2 middleware, and is containerized using Docker.
We load an additional container on each robot that implements \textsc{D-D4orm} as a standalone ROS~2 node in C++.
This container interfaces to the robot's controller via \texttt{localhost} (to execute trajectories), and to other robots via the Jetson's default Wi-Fi modules.
Our setting exploits multi-threading on the ARMv8 CPU.

\cref{fig:overview}(f) shows a timelapse-snapshot from a distributed on-robot execution using six RoboMaster platforms.
The implementation showcases a ``lifelong'' asynchronous mission update, where robots are continuously and asynchronously given new (random) targets in the workspace as soon as they complete the current mission.
This creates a dynamism that is closer to real-world settings (e.g., in warehouses).

\cref{fig:real-world-dist} shows a detailed process visualization for the timeline of distributed computation and message exchange.
The team is comprised of six robots that also act as computing nodes.
A new problem is assigned to the team at time $t{=}0$, and the first iteration lasts for $\approx$\SI{600}{ms}, at the end of which, all nodes exchange their candidate solutions.
The best candidate is visualized in the top row.
Note that a viable solution is found in the first iteration; however, to illustrate the \textit{anytime} nature of \textsc{D4orm}, we let the agents execute the full number of iterations (${=}5$ here).
For each successive iteration, every agent picks a random candidate solution, shown as a solid black connector, from the top-$k$ subset.
At the end of the 5th iteration, the solution from Robot-5 is ranked the highest, and is the one that all robots execute.
The process is successful in spite of variable latency and a lack of message synchronization in the inter-robot communication (over stock ROS~2 DDS middleware).
However, consequently, the agents spend ${\approx}\SI{720}{ms}$ not performing any computation (averaged across agents and then summed), which corresponds to ${\approx}20\%$ of the interval in this illustration.
These statistics are typical of stock DDS settings over general-purpose infrastructure Wi-Fi networks, and may be improved using customized protocols that are tuned appropriately for the task.

\section{Conclusion}
\label{sec:conclusion}
This paper studied a multi-robot trajectory optimization framework with diffusion denoising.
Given robot dynamics and a reward function, \textsc{D4orm} (an offline, coupled planning framework) iteratively applies a deformation vector to the current solution candidate of a joint trajectory for the team.
This process is guided by a Monte Carlo gradient approximation.
Despite the nonconvexity, multimodality, and high-dimensionality of deconfliction problems, both our theoretical and empirical analyses suggest that this simple process successfully retrieves collision-free trajectories for the entire team within reasonable deadlines.
We then use vanilla \textsc{D4orm} as a building block to develop advanced variants, such as a decoupled planner \textsc{D4orm-D} to achieve scalability, the online planning scheme \textsc{D4orm-C2F}, and the distributed version \textsc{D-D4orm} for resource-constrained environments.
Taken together, the results of this study provide a toolkit for solving various types of multi-robot trajectory planning problems.
This opens up the possibility of diffusion-based sampling optimization methods for multi-robot trajectory planning, an area that has been largely under-explored.

While the \textsc{D4orm} series provides powerful and convenient solutions, we acknowledge several technical limitations.
For example, \textsc{D4orm} itself remains a local collision-avoidance planner and is not designed to solve trajectory planning in obstacle-rich workspaces such as maze-like environments.
In such cases, integration with discrete search, especially multi-agent pathfinding methods~\cite{stern2019def}, would yield more practical approaches.
Another limitation is the assumption of fast collision checking to compute the reward (e.g., using sphere-like robot shapes).
With non-convex robot shapes, collision checking would become a bottleneck and slow down the Monte Carlo rollout, thereby degrading the performance of \textsc{D4orm}.
This will be an important consideration when adapting \textsc{D4orm} series to more advanced dynamics, such as manipulators.
Nevertheless, we envision that denoising optimization will inspire a new class of multi-robot planning algorithms, enabling scalable, flexible, and robust frameworks that extend beyond the current \textsc{D4orm} styles.

\appendices
\section{Discovering Multimodal Solutions}
\label{sec:appdx:multi-modality}
\input{figures/d4orm_multi_modal}
\newcommand{\entry}[3]{
  \node[] at (\xsize * #1, #2) {
    \includegraphics[width=0.3\linewidth]{images/baseline_cmp_appendix/#3}
  };
}

\begin{figure}
\vspace{1pt}
\centering
\begin{tikzpicture}
  \scriptsize
  \tikzmath{
    \xsize = 2.75;
    \ya = 0;
  }

  \node[] at (0.1 + \xsize * 0, 1.2) {Differential Drive};
  \node[] at (0.1 + \xsize * 1, 1.2) {2D Holonomic};
  \node[] at (0.1 + \xsize * 2, 1.2) {3D Holonomic};

  \entry{0}{\ya}{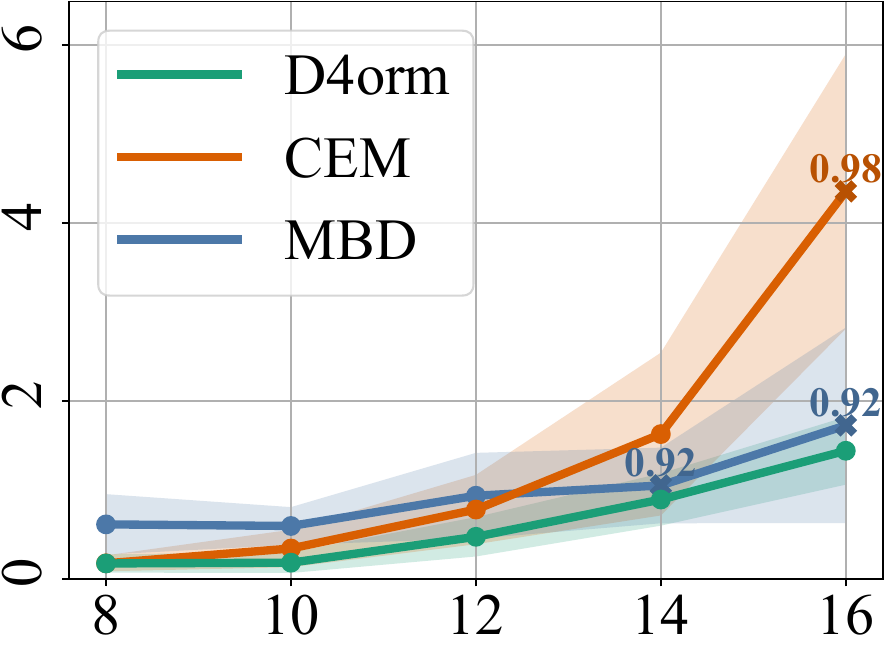}
  \entry{1}{\ya}{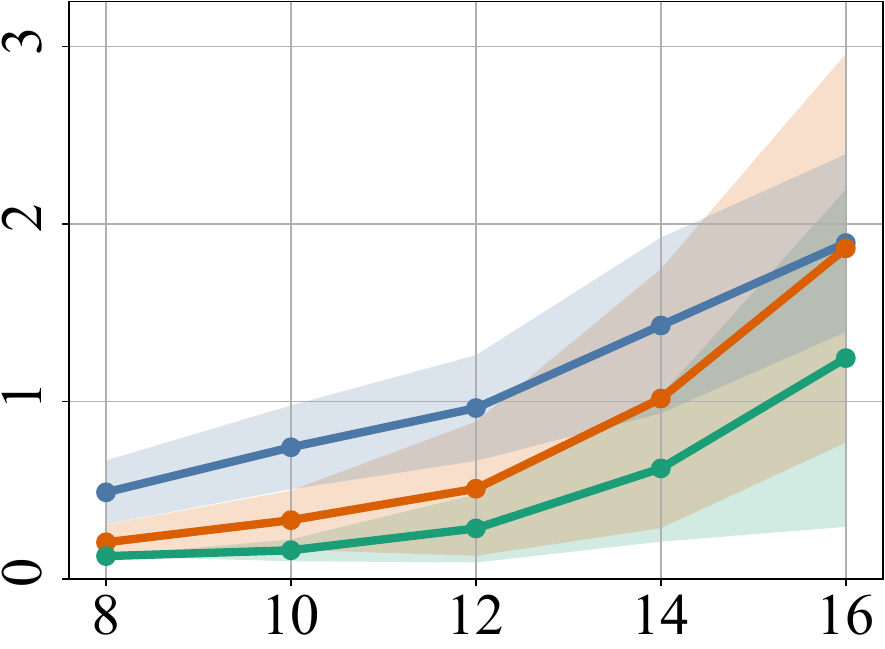}
  \entry{2}{\ya}{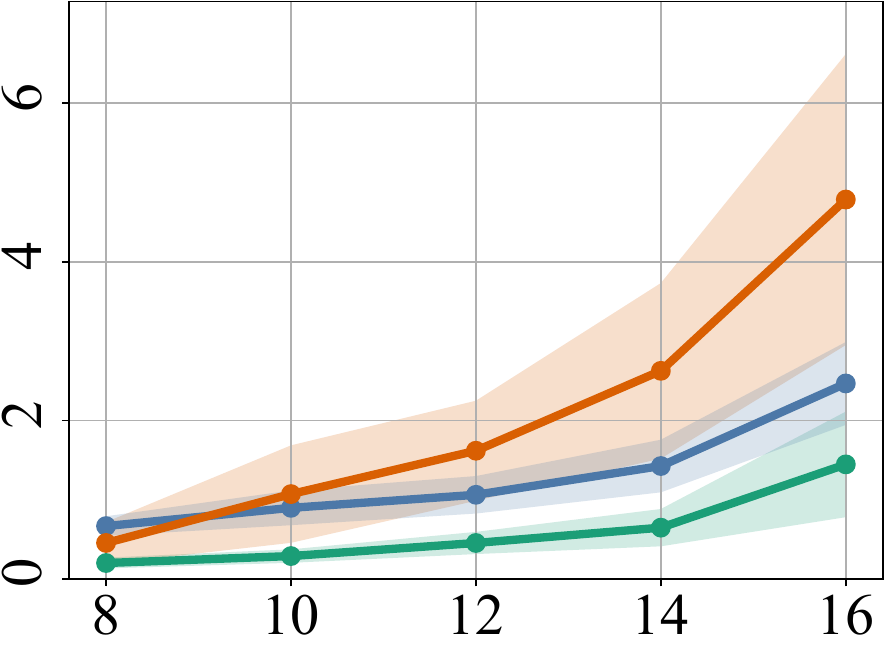}

  \node[rotate=90] at (-\xsize * 0.5 - 0.2, \ya)
    {$\leftarrow$ runtime [s]};

  \node[] at (\xsize * 1, \ya - 1.2)
    {\#robots ($\times$: exclude failed runs, w/success rates; $\bullet$: all runs succeeded)};

\end{tikzpicture}
\caption{
\textsc{D4orm}'s performance against CEM and MBD. 
The number of denoising steps $N$ is set to 100 for \textsc{D4orm} and 10,000 for MBD. 
For a fair comparison, all methods evaluate whether the current trajectory is successful every 100 steps.
}
\label{fig:baseline_cmp_appendix}
\end{figure}

In \cref{sec:d4orm-analysis:theory}, we motivated the importance of capturing the \emph{multimodality} of valid multi-robot trajectories.
Since \textsc{D4orm} and its variants utilize a sampling-based diffusion-denoising scheme, we hypothesized that they can discover a diverse set of candidate solutions rather than converging to a limited number of modes.
Here, we examine this capability through case studies comparing \textsc{D4orm} against numerical trajectory optimization (i.e., a direct collocation) with \texttt{IPOPT} used in \cref{sec:d4orm-analysis:evals:ipopt}.

In \cref{fig:d4orm_multi_modal}, we empirically investigate this ability using sample problems with 2D holonomic robots.
The left panel illustrates a relatively simple multimodal setting, where two robots and their goals are at antipodal points of a circle, the center of which contains one obstacle.
As illustrated, this problem admits six distinct solution classes.
We run \textsc{D4orm} and \texttt{IPOPT} 100 times using different seeds and initial guesses, and project the resultant trajectory features (2D positions) onto two principal components using PCA.
The visualization clearly reveals these six clusters, and both methods successfully recover all six solution classes.

In more challenging scenarios, however, \textsc{D4orm}'s capacity to uncover diverse, multimodal solutions becomes particularly apparent.
\cref{fig:d4orm_multi_modal} (right) shows the same trajectory-feature plot for a problem that involves eight robots.
The increased complexity of this setting gives rise to a substantially larger set of feasible solutions, and \textsc{D4orm} is able to discover a broader range of them compared to \texttt{IPOPT}.

\section{Comparison with CEM and MBD}

In addition to the comparison with MPPI and i-MBD presented in \cref{sec:d4orm-analysis:eval}, we further compare \textsc{D4orm} against CEM~\cite{botev2013cross} and vanilla MBD (single-iteration)~\cite{pan2024model}. 
For CEM, the elite fraction is set to 5\%. For MBD, we use a large number of denoising steps ($N{=}10{,}000$) to ensure sufficient steps for all test cases, while \textsc{D4orm} keeps $N{=}100$ for each iteration and runs iteratively. All methods use 1,024 samples per step and evaluate trajectory feasibility every 100 steps.

As shown in \cref{fig:baseline_cmp_appendix}, \textsc{D4orm} achieves higher success rates and lower runtimes than both methods. MBD is particularly slow for smaller robot teams, indicating that a fixed number of denoising steps does not generalize well across problems of varying complexity. This stems from its noise schedule, whose covariance depends on the fraction of completed steps rather than their absolute number. In contrast, \textsc{D4orm}-family methods use iterative deformation to adapt more efficiently to varying problem complexity.

\section{Additional Sensitivity Analyses}

\subsection{Noise Schedule}
\input{figures/d4orm_parameters}

\cref{fig:d4orm_parameters} shows the impact of a varying noise schedule.
Following the standard diffusion model notation, we express the schedule in terms of $\beta_t$, with $\alpha_t = 1 - \beta_t$ and $\bar{\alpha}_i = \prod_{j = 1}^i \alpha_j$.
A larger $\beta_1$ injects more noise early in the forward process (late in reverse denoising), whereas a larger $\beta_T$ injects more noise late in the forward process (early in reverse denoising).
The variation in performance is generally small, indicating that the iterative refinement is able to adapt to different noise schedules by expending more iterations.
The difference is most observable in the differential drive case, where starting with a high $\beta_T$ is likely to produce infeasible trajectories.

\subsection{Coarse Trajectory for \textsc{D4orm-C2F}}
{
\begin{table}[t!]
\centering
\caption{Sensitivity of \textsc{D4orm-C2F} to the coarse-guidance parameters under 2D Holo 4 Obs, using 16 robots.}
\label{tab:d4orm_c2f_parameter_tab}
\renewcommand{\arraystretch}{1.0}
\setlength{\tabcolsep}{1mm}
\begin{tabular}{cc|ccc}
\toprule
$f\coarse$ & $o\coarse$ & runtime [\si{\second}] & average travel time [\si{\second}] & makespan [\si{\second}] \\
\midrule
1 & 1  & 0.75 & 6.23 & 7.23 \\
\midrule
2 & 1  & 0.37 (-0.38) & 6.27 (+0.04) & 7.31 (+0.08) \\
3 & 1  & 0.32 (-0.43) & 6.35 (+0.12) & 7.52 (+0.29) \\
4 & 1  & 0.23 (-0.52) & 6.42 (+0.19) & 7.62 (+0.39) \\
\midrule
1 & 5  & 0.57 (-0.18) & 6.31 (+0.08) & 7.49 (+0.26) \\
1 & 10 & 0.43 (-0.32) & 6.36 (+0.13) & 7.58 (+0.35) \\
\bottomrule
\end{tabular}

\vspace{1mm}
\parbox{\columnwidth}{\footnotesize
Parenthesized values indicate absolute differences from the base setting ($f\coarse{=}1$ and $o\coarse{=}1$).
}
\end{table}
}

\cref{tab:d4orm_c2f_parameter_tab} evaluates the impact of the hyperparameters in \textsc{D4orm-C2F} using 2D holonomic scenarios with four obstacles. 
We focus on the time-scaling factor $f\coarse$ and the collision-check interval $o\coarse$, both of which determine the granularity of the generated coarse trajectory.
For both parameters, larger values indicate coarser guidance, resulting in faster solution discovery, as shown in the result.
However, excessively coarse guidance also degrades solution quality, leading to increased average travel time of each robot and makespan. 
This highlights a trade-off between computational overhead and execution efficiency: coarser trajectories enable faster planning but yield less effective guidance, resulting in longer mission completion times. 
Moreover, the runtime improvements exhibit diminishing returns as $f\coarse$ and $o\coarse$ continue to increase. 
These results suggest that moderately coarse guidance provides the most favorable balance between planning overhead and solution quality.

\section*{Acknowledgment}
The authors would like to thank the anonymous reviewers and the editor for their insightful comments, and Lorenzo Magnino for his assistance with the robot deployment.

\bibliographystyle{IEEEtran}
\bibliography{sty/ref-macro-short,ref}

\begin{IEEEbiography}
[{\includegraphics[
    width=1in,
    height=1.25in,
]{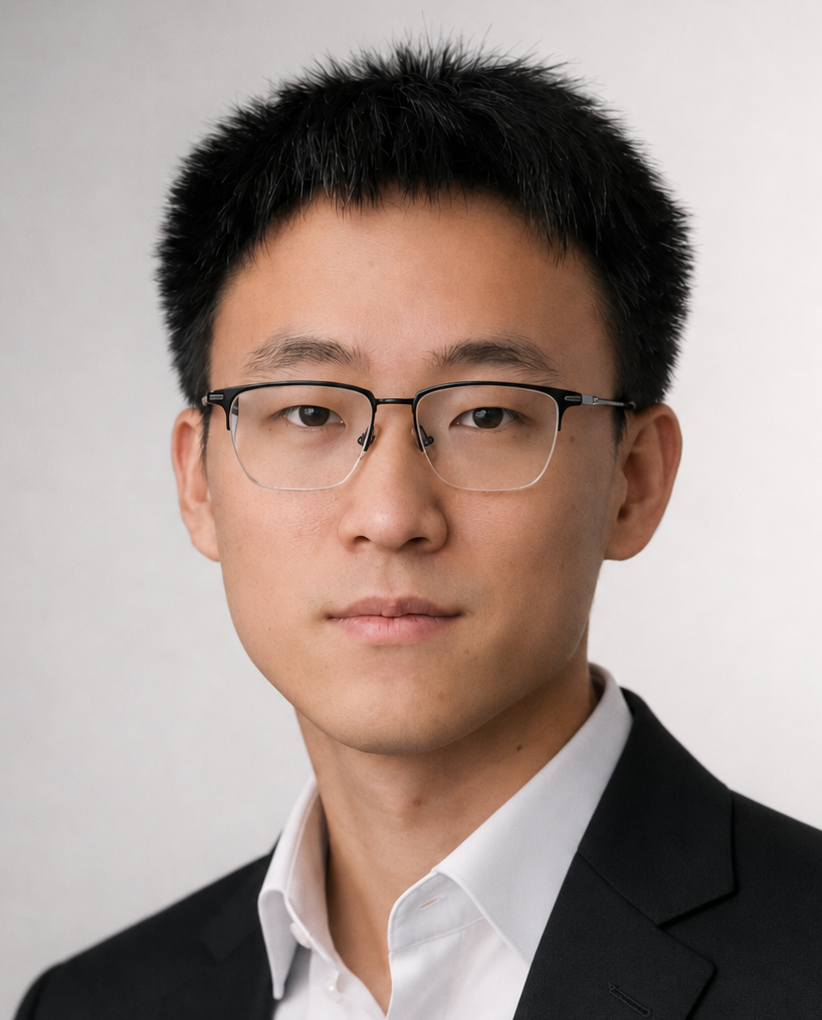}}]
{Yuhao Zhang} (Graduate Student Member, IEEE) received the B.Sc. and M.Phil. degrees in computer science from the University of Manchester in 2024 and the University of Cambridge in 2025, respectively. He is currently pursuing the Ph.D. degree in engineering science at the Oxford Robotics Institute, University of Oxford. His research focuses on robot perception and planning, with particular interests in multi-robot trajectory planning, simultaneous localization and mapping, and spatial intelligence.
\end{IEEEbiography}

\begin{IEEEbiography}
[{\includegraphics[
    width=1in,
    height=1.25in,
]{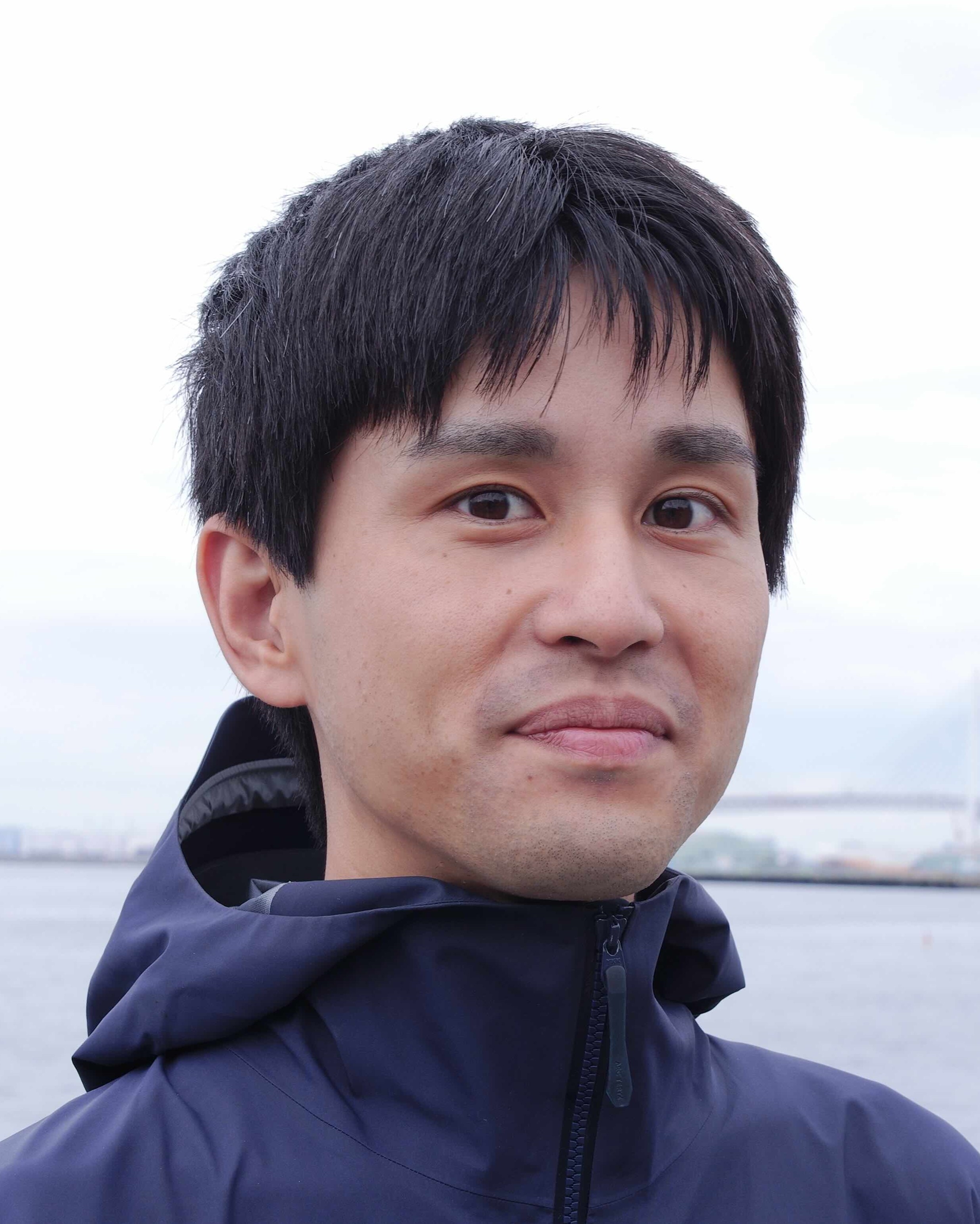}}]
{Keisuke Okumura} (Member, IEEE) received the Ph.D. degree in computer science from Tokyo Institute of Technology, Tokyo, Japan, in 2023.
He was subsequently a Visiting Postdoctoral Scholar at the University of Cambridge, Cambridge, U.K., until the end of 2025.
Since 2023, he has been a Researcher at the National Institute of Advanced Industrial Science and Technology (AIST), Tokyo,~Japan.
His research focuses on developing scalable AI methods for collective automation, particularly at the intersection of multi-agent planning, heuristic search, learning, and multi-robot coordination. 
His honors include the ICAPS 2022 Best Student Paper Award, Tokyo Tech’s Ph.D. Dissertation Award (2024), and Funai Information Technology Award for Young Researchers (2026).
\end{IEEEbiography}

\begin{IEEEbiography}
[{\includegraphics[
    width=1in,
    height=1.25in,
]{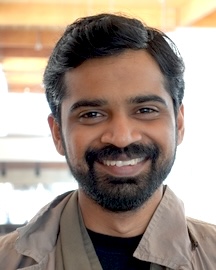}}]
{Ajay Shankar} (Member, IEEE) received a Ph.D. in Computer Science (Robotics) from the University of Nebraska-Lincoln, USA.
He subsequently worked as a post-doctoral Research Associate and an Assistant Research Professor in the Department of Computer Science and Technology at the University of Cambridge, U.K.
Until mid-2026, he also held a position there as a visiting research scholar.
His research explores efficient, distributed and decentralized methods for multi-robot control, planning, and learning, with a particular interest in applied settings in field and experimental robotics.
\end{IEEEbiography}

\begin{IEEEbiography}
[{\includegraphics[
    width=1in,
    height=1.25in,
]{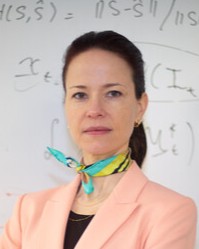}}]
{Amanda Prorok} (Senior Member, IEEE) received the Ph.D. degree from EPFL - Swiss Federal Technology Institute of Lausanne, Lausanne, Switzerland, in 2013.
She is currently a Professor of Collective Intelligence and Robotics with the Department of Computer Science and Technology, University of Cambridge, Cambridge, U.K., and a Fellow of Pembroke College, Cambridge, U.K. 
In her work, she pioneered differentiable communications methods for multi-agent systems, with applications to multi-robot perception and control. Amanda has given invited keynotes at TEDx and IEEE ICRA and has received numerous research awards, including a prestigious ERC Starting Grant. Amanda is an IEEE Senior Member, serves as Senior Editor for IEEE Transactions on Robotics (T-RO) and chaired the 2021 IEEE International Symposium on Multi-Robot and Multi-Agent Systems. Her Ph.D. thesis received the Asea Brown Boveri (ABB) prize for the best thesis in Computer Science at EPFL. 
\end{IEEEbiography}

\vfill

\end{document}